\documentclass[runningheads]{llncs}

\usepackage{eccv}

\usepackage{eccvabbrv}

\usepackage{graphicx}

\usepackage[accsupp]{axessibility}  

\usepackage[dvipsnames]{xcolor}

\definecolor{turquoise}{cmyk}{0.65,0,0.1,0.3}
\definecolor{purple}{rgb}{0.65,0,0.65}
\definecolor{dark_green}{rgb}{0, 0.5, 0}
\definecolor{orange}{rgb}{0.8, 0.6, 0.2}
\definecolor{dark_orange}{rgb}{0.7, 0.6, 0.3}
\definecolor{red}{rgb}{0.8, 0.2, 0.2}
\definecolor{darkred}{rgb}{0.6, 0.1, 0.05}
\definecolor{blueish}{rgb}{0.0, 0.3, .6}
\definecolor{light_gray}{rgb}{0.7, 0.7, .7}
\definecolor{pink}{rgb}{1, 0, 1}
\definecolor{cyan}{rgb}{0., 1, 1}

\definecolor{checkyes}{rgb}{0.7, 1.0, 0.7}
\definecolor{crossno}{rgb}{1.0, 0.7, 0.7}

\definecolor{tabbestcolor}{rgb}{0.785, 0.851, 0.969}

\definecolor{mydarkblue}{rgb}{0,0.08,0.55}

\renewcommand{\paragraph}[1]{\vspace{.2em}\noindent\textbf{#1}.}

\usepackage{amsmath}
\usepackage{pifont}
\usepackage{graphicx}
\usepackage{subcaption}
\usepackage{amssymb}
\usepackage{booktabs}
\usepackage{currfile}

\usepackage{enumitem}  
\usepackage{xcolor}  
\usepackage{color}
\usepackage{colortbl}
\usepackage{microtype}
\usepackage{nicefrac}
\usepackage{siunitx}
\usepackage[hang,flushmargin]{footmisc}
\usepackage{xfrac}
\usepackage[section]{placeins}
\usepackage{etoolbox}
\usepackage{bbm}

\newcommand{\cmark}{\textcolor{green}{\text{\ding{51}}}}
\newcommand{\xmark}{\textcolor{red}{\text{\ding{55}}}}
\newcommand{\redcmark}{\textcolor{red}{\text{\ding{51}}}}
\newcommand{\greenxmark}{\textcolor{green}{\text{\ding{55}}}}
\usepackage{multirow}
\usepackage{gensymb}
\usepackage{arydshln}
\usepackage{tikz}
\usepackage{times}
\usepackage{epsfig}
\usepackage{pgfplots} 
\usepackage{pgfplotstable} 
\pgfplotsset{compat=newest} 

\usepackage[pagebackref,breaklinks,colorlinks,citecolor=eccvblue]{hyperref}

\usepackage{orcidlink}

\begin{document}

\title{OmniNOCS: A unified NOCS dataset and model for 3D lifting of 2D objects}

\titlerunning{OmniNOCS}

\author{Akshay Krishnan\inst{1,2}\orcidlink{0009-0005-0974-6882} \and
Abhijit Kundu\inst{1}\orcidlink{0000-0002-7001-477X} \and
Kevis-Kokitsi Maninis\inst{1}\orcidlink{0000-0003-3776-0049}
\and
James Hays\inst{2}\orcidlink{0000-0001-7016-4252}
\and
Matthew Brown\inst{1}\orcidlink{0009-0004-9749-7363}
}

\authorrunning{A.~Krishnan et al.}

\institute{Google Research\footnote{Now at Google DeepMind} \and Georgia Institute of Technology}

\maketitle

\def\boxmethod{Cubeformer}
\def\nocsmethod{NOCSformer}
\def\cubercnn{Cube R-CNN}
\def\numimages{380k}
\def\numclasses{97}
\def\numsources{10}

\begin{figure}
\centering
\vspace*{-3em}  
\captionsetup{type=figure}
\includegraphics[width=.98\textwidth]{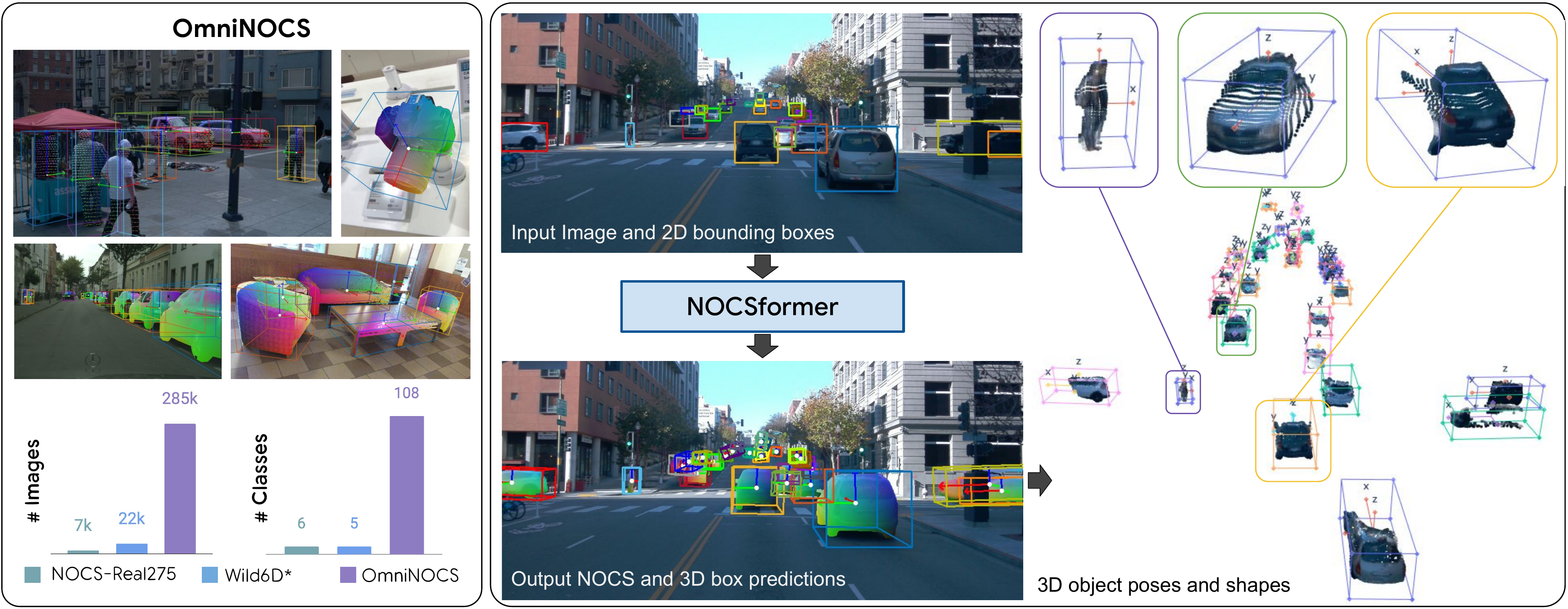}
\vspace*{-5px}  
\captionof{figure}{\small{We introduce \textbf{OmniNOCS}, a large-scale dataset with Normalized Object Coordinates (NOCS), instance masks, and 3D box annotations for objects across several classes, domains, and cameras. We also propose \textbf{\nocsmethod{}}, a model trained on OmniNOCS that lifts each 2D object bounding box in an image to its corresponding 3D oriented box (pose) and 3D pointcloud (shape).}}
\label{fig:teaser}
\end{figure}
\vspace*{-1em}

\vspace{-10mm}
\begin{abstract}
We propose OmniNOCS, a large-scale monocular dataset with 3D Normalized Object Coordinate Space (NOCS) maps, object masks, and 3D bounding box annotations for indoor and outdoor scenes. OmniNOCS has 20 times more object classes and 200 times more instances than existing NOCS datasets (NOCS-Real275, Wild6D). 
We use OmniNOCS to train a novel, transformer-based monocular NOCS prediction model (NOCSformer) that can predict accurate NOCS, instance masks and poses from 2D object detections across diverse classes. It is the first NOCS model that can generalize to a broad range of classes when prompted with 2D boxes. 
We evaluate our model on the task of 3D oriented bounding box prediction, where it achieves comparable results to state-of-the-art 3D detection methods such as \cubercnn{}. Unlike other 3D detection methods, our model also provides detailed and accurate 3D object shape and segmentation. We propose a novel benchmark for the task of NOCS prediction based on OmniNOCS, which we hope will serve as a useful baseline for future work in this area. Our dataset and code will be at the project website: \url{https://omninocs.github.io}.
\end{abstract}

\vspace*{-1em}
\vspace{-8mm}
\section{Introduction}
\label{sec:intro}
\vspace{-1mm}

Predicting the 6 Degree-of-Freedom (6DoF) pose and shape of objects from images is a crucial problem in 3D scene understanding. Robots need to understand the location, shape, and orientation of various objects to grasp and interact with them. Self-driving vehicles need to understand the location and heading of vehicles, pedestrians, and other objects on the road. In particular, the 3D orientation of these objects is crucial to predict their future behavior. It is also important in AR/VR applications, as it allows users to interact with objects in meaningful ways. Predicting object shape and pose from monocular images is also a prerequisite to initialize panoptic 3D neural scene graph representations \cite{KunduCVPR2022PNF,nie2020total3dunderstanding} and for methods that track objects in videos \cite{rajasegaran2022tracking}. Most of these applications require approaches that generalize across a wide range of object classes, environments, and camera models.

The problem of localizing 3D objects has been extensively studied through the lens of monocular 3D object detection or 6DoF pose estimation. The most commonly used approach is to represent objects as 3D cuboids, and train a model to regress the cuboid parameters from a 2D RoI of the object \cite{3DRCNN_CVPR2018, brazil2023omni3d, wang2021fcos3d, brazil2019m3d, xiang2017posecnn, rad2017bb8}. An alternative method involves predicting corresponding 3D points in an object coordinate space followed by pose estimation using 3D-2D alignment~\cite{chen2022epro,liu2021autoshape,Wang_2019_CVPR,goodwin2022zero,monorun2021}. However, all these methods are limited to narrow datasets collected on a single camera and context with typically very few object classes. Existing models are also trained separately for every dataset, preventing them from being used more widely.

Recent work \cite{brazil2023omni3d} takes a step towards scalability by introducing the Omni3D dataset, which aggregates 3D detection datasets from different domains, and trains a \cubercnn{} model to regress 3D bounding boxes for 50 classes. However, Omni3D notably lacks consistently oriented, object-centric boxes, as the ground truth orientations are not canonical for object classes. \cubercnn{} \cite{brazil2023omni3d} instead uses a Chamfer distance loss which is invariant to the predicted 3D cuboid orientation. This causes it to predict inconsistent object orientations, for example, often flipping the orientations of cars by 180\degree. Further, Omni3D is a detection dataset that does not provide object shape information. 

Our work aims to overcome the above shortcomings, by providing a new large-scale dataset with consistent object-centric ground truth boxes along with detailed shape. We argue for the use of Normalized Object Coordinate Space (NOCS) as proposed in \cite{Wang_2019_CVPR} as a 3D object shape representation. NOCS represents both \textit{the canonical orientation and the shape} of the visible surface of the object, properties that are essential for real-world applications such as self-driving and robotics. It can also be used to estimate the 3D bounding box of the object. However, all existing work on NOCS \cite{Wang_2019_CVPR, li2019cdpn, wang2021gdr, irshad2022centersnap} only train small models on small datasets with fewer than 10 classes. We address the lack of diverse NOCS datasets by proposing a large-scale monocular NOCS dataset, \textbf{OmniNOCS} that has NOCS annotations, instance segmentation, and canonically oriented bounding boxes for \numclasses{} object classes, containing \numimages{} images from \numsources{} different data sources, making it the largest and most diverse NOCS dataset currently available. OmniNOCS includes all of the data from Omni3D (KITTI \cite{Geiger2012CVPR}, nuScenes \cite{caesar2020nuscenes}, ARKitScenes \cite{dehghan2021arkitscenes}, Objectron \cite{objectron2021}, Hypersim \cite{hypersim}, SUN-RGBD \cite{sunrgbd}), with the addition of Cityscapes \cite{cordts2016cityscapes}, virtual KITTI \cite{cabon2020virtual}, NOCS-Real275 \cite{Wang_2019_CVPR}, and the Waymo Open Dataset \cite{waymodataset}.

Our work also introduces a new model, which we term ``\nocsmethod{}'', that predicts NOCS coordinates and oriented 3D bounding boxes from monocular images and 2D detections for all the classes in OmniNOCS. \nocsmethod{} leverages large self-supervised pretrained ViTs \cite{dosovitskiy2020image, oquab2023dinov2}. It does not use any class-specific heads or parameters. This enables it scale to large vocabularies, and share information across semantically similar object classes. In contrast, existing NOCS models use small NOCS heads with class-specific parameters that significantly limit their performance on large datasets like OmniNOCS. Apart from the NOCS predictions, \nocsmethod{} also predicts the 3D size of the object, and 3D orientation using a learned PnP head. This allows \nocsmethod{} to predict 3D oriented bounding boxes and object point cloud in metric scale for diverse object categories. Our experiments evaluate the quality of NOCS and bounding boxes predicted by \nocsmethod{} in comparison to existing NOCS prediction or 3D detection models. We find that training on OmniNOCS allows \nocsmethod{} to generalize to unseen datasets, even outperforming baselines trained on the target dataset in NOCS prediction accuracy. 

In summary, our contributions are:
\vspace{-5px}
\begin{itemize}
\item \textbf{The OmniNOCS dataset}: A new dataset containing Normalised Object Coordinates for \numimages{} images in \numclasses{} categories, an order of magnitude larger on both counts than existing NOCS datasets.
\item \textbf{\nocsmethod{}}:  A novel transformer-based architecture for predicting object NOCS, mask, and size from input 2D boxes, utilizing pre-trained self-supervised ViT backbones. \nocsmethod{} is the first NOCS model to generalize to vocabularies with 90+ classes and to unseen datasets, including images from internet collections. 
\item \textbf{OmniNOCS benchmark}: an evaluation framework with metrics for directly comparing different NOCS prediction algorithms on OmniNOCS, with baselines established via NOCSFormer. 
\end{itemize}

\section{Related work}
\label{sec:relatedwork}




The task of predicting 3D object pose has been studied both in the context of camera/object pose estimation (6DoF) and 3D bounding box estimation (6DoF pose + 3DoF size).
A further distinction can be made between methods that regress directly to object pose / bounding box parameters, and methods that compute per pixel coordinates or depth as an initial step. We review each of these paradigms in the sections below.


\subsection{3D localization by regressing bounding boxes}

Direct approaches to 3D object detection and pose estimation involve networks that output rotation, translation and scale parameters directly. Several works have explored different parameterizations in this setting, e.g., PoseCNN~\cite{xiang2017posecnn} uses regression of translation via centre direction + distance maps (which enable detection even under occlusion), and quaternion representation of rotation. BB8~\cite{rad2017bb8} uses segmentation followed by a CNN to regress to 2D projections of the 8 $\times$ 3D bounding box corners. Multi-view monocular approaches have also been proposed, e.g., DETR3D~\cite{wang2022detr3d} which uses DETR-style attention to reason about object interactions. Several techniques also make use of existing 2D bounding box predictors, either as an input to a 3D lifting approach~\cite{manhardt2019roi}, or as a constraint on 3D box predictions~\cite{mousavian20173d}.

Another group of works focuses on predicting the alignment of 3D CAD models within various modalities: images~\cite{kuo2020mask2cad,gumeli2022roca}, videos~\cite{li2021odam,maninis2022vid2cad}, or 3D scans~\cite{avetisyan2019scan2cad,avetisyan2019end}.  These approaches typically determine the object's 3D translation, rotation, and size (9DoF) and find a CAD model with a similar visual appearance.

\subsection{3D localization from model-to-image alignment}

An alternative approach to 3D localization is to first predict correspondence, either between views or to a normalized space, and then reason over redundant correspondences to establish pose. This has been done with both sparse and dense correspondences, e.g., AutoShape~\cite{liu2021autoshape} makes use of sparse 2D to 3D correspondences, with a learned shape model and sparse 2D keypoints. \cite{goodwin2022zero} uses pairwise semantic correspondence from ViT, to find pose between a reference image and a sequence of targets.

Several approaches use dense correspondences and the idea of Normalised Object Coordinate Spaces, or NOCS. In the original NOCS work, Wang et al.~\cite{Wang_2019_CVPR} use RGB-D data and 3D-3D correspondences for pose estimation. 
Follow-up work \cite{monorun2021} estimates pose using PnP variants on 2D-3D correspondences. While NOCS \cite{Wang_2019_CVPR} used the average spatial dimensions of the object category to define the object frame, \cite{wen2021catgrasp} showed that an instance-specific NOCS coordinate frame can be used alongside a predicted instance size for the same purpose. Other methods combine coordinate regression and direct approaches, e.g., ~\cite{wang2021gdr}, which uses direct regression based on dense correspondences.  

Similar to our work, MonoRUn~\cite{monorun2021}  lifts 2D detections to 3D object coordinates without explicit class supervision. They also use a novel self-supervised coordinate regression training loss which obviates the need for detailed 3D ground truth. However, their representation does not include masks and therefore does not provide explicit object shape information. Their evaluation is also limited to 3 classes (car, pedestrian, cyclist) on the KITTI-Object test set. 
M3D-RPN~\cite{brazil2019m3d} also has a single 3D head for multiple classes, jointly generating 2D and 3D bounding box predictions, though without shape information, and similarly limited to car, pedestrian and cyclist classes on KITTI.

\subsection{Monocular object localization / pose estimation datasets}


For 3D object detection from monocular RGB images, most existing works use a small number of classes on a single dataset. \cubercnn{}/Omni3D~\cite{brazil2023omni3d} contribute towards creating a general purpose monocular 3D object detector. \cubercnn{} performs well over six 3D datasets: KITTI, SUN RGB-D, ARKitScenes, Objectron, nuScenes and Hypersim. However, the method is class specific, with \cubercnn{} trained only on 50 classes, and unable to work in the open class setting.

\cubercnn{} also uses a Chamfer loss on the predicted 3D box corners to deal with the inconsistent coordinate frame annotations in the underlying Omni3D datasets. The lack of direct orientation supervision results in inconsistent orientation predictions (for example, the positive x axis may point to the front or the back of the car for different instances). This orientation inconsistency is problematic for methods that seek to build detailed 3D models of object categories as a downstream task~\cite{KunduCVPR2022PNF,nie2020total3dunderstanding}.

Other works are not limited to 3D boxes, and directly provide object shapes. These shapes come in the form of annotated point clouds (e.g., ScanNet~\cite{dai2017scannet}), or aligned CAD models (e.g., Scan2CAD~\cite{avetisyan2019scan2cad}, Pix3D~\cite{sun2018pix3d}, CAD-Estate~\cite{maninis2023cad}). Object point clouds are created by labeling points on scanned 3D scenes, a process that doesn't scale well to large datasets.
Semi-automatically aligning CAD models is scale-able to a certain extent, but since it relies on retrieving existing models, the resulting shapes are rarely accurate, and the alignments are sensitive to deformations and movable parts.
In contrast to these works, we propose a large-scale dataset of many different categories.

\section{OmniNOCS dataset}
\label{sec:omninocs-dataset}

\begin{table}[t]
\centering
\begin{minipage}{.7\linewidth}
\resizebox{1\linewidth}{!}{
\begin{tabular}{ l | c | c | c c c} 
 \hline
 Dataset & NOCS GT & Real & \#Images & \#Classes & \#Instances  \\ 
 \hline
 CAMERA25 \cite{Wang_2019_CVPR} & \cmark & \xmark & 300k & 6 & 184  \\ 
 NOCS-Real275 \cite{Wang_2019_CVPR}  & \cmark & \cmark & 7k & 6 & 24 \\ 
 Wild6D \cite{fucategory} & \xmark & \cmark & 1M & 5 & 1.8K  \\ 
 \hline
 \textbf{OmniNOCS} & \cmark & \cmark  & \numimages{} & \numclasses{} & > 450k \\ 
 \hline
\end{tabular}
}
\end{minipage}
\vspace{1mm}
\caption{\small{\textbf{Comparison to other NOCS datasets}: Other existing NOCS datasets are limited in number of classes and instances. Note that Wild6D has 1M images, but these are used for self-supervised training, as it does not include NOCS ground truth (GT).}}
\label{table:dataset-comparison}
\vspace{-4mm}
\end{table}


\subsection{Data statistics}

We create OmniNOCS, a first-of-its-kind \emph{large and diverse} NOCS dataset comprising data from several classes and domains.  OmniNOCS uses data from self-driving scenes \cite{Geiger2012CVPR, cordts2016cityscapes, waymodataset, cabon2020virtual}, indoor scenes \cite{hypersim, dehghan2021arkitscenes, Wang_2019_CVPR, sunrgbd}, and object-centric videos \cite{objectron2021}. Each of these sources use cameras with different parameters, ranging from phone cameras to wide FOV cameras mounted on self-driving cars. The number of object instances per image also varies widely, from single-object images~\cite{objectron2021} to images with hundreds of objects~\cite{waymodataset, hypersim}. 

OmniNOCS has \numclasses{} object categories across \numimages{} images. This by far exceeds the diversity of existing NOCS datasets (with less than 6 categories as shown in Table \ref{table:dataset-comparison}). The number of instances exceeds that of other NOCS datasets by more than 2 orders of magnitude. It also contains more images with NOCS annotations (\numimages{}). Note that while Wild6D has 1M images, these do not have ground truth NOCS or pose annotations, and includes every frame of object-centric videos. OmniNOCS is also more diverse than the most diverse 3D detection dataset (Omni3D \cite{brazil2023omni3d} with 98 categories). 307k images in OmniNOCS are from real world scenes and 73k are synthetic. 107k of the images in OmniNOCS are from outdoor scenes, 102k images are from complex indoor environments, and 140k are from object-centric videos. 

\subsection{Data preparation}

\begin{figure*}[tb]
    \centering
    \includegraphics[width=1\linewidth]{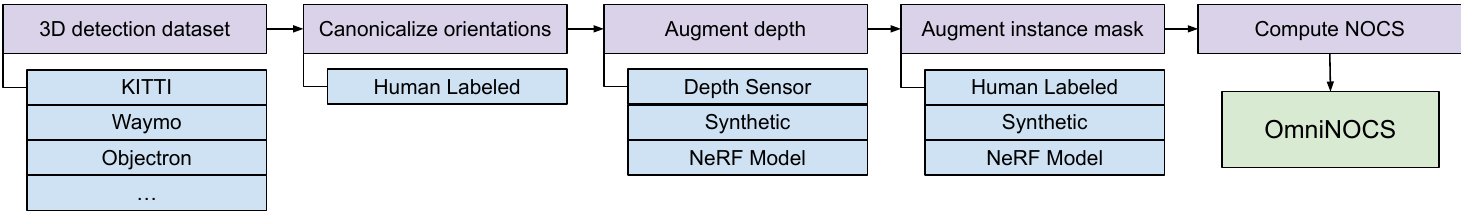}
    \vspace{-4mm}
    \caption
    {
    \small{
    \textbf{Preparation of the OmniNOCS dataset:} First, we ensure that object orientations are canonical across different all subsets of OmniNOCS. Next, we re-compute depth for datasets where depth is not available or is too noisy. Finally, we annotate objects with (pseudo) instance mask labels where ground truth masks are not available.
    }
    }
\label{fig:dataset_prep}
\end{figure*}

Computing the NOCS for objects in an image requires canonically oriented 3D object bounding boxes, depth, and instance segmentations. Since the datasets we use were originally only proposed for monocular 3D detection, many of them lack dense depth or instance segmentation annotations (highlighted entries in Table \ref{table:nocs_datasets}). Additionally, NOCS also requires all instances within a category to have consistent orientation annotations. While this is certainly not true \textit{across} different 3D detection datasets, it is also not true \textit{within} some large synthetic datasets such as Hypersim. OmniNOCS aggregates several datasets, with the addition of 1) new depth estimated via Mip-NeRF reconstructions, 2) additional instance masks via segmentation models and human labels, 3) manual labelling of coordinate axes to give consistent object-centric coordinate frames. This enables NOCS to be computed for every image in the dataset.  This multi-stage process for OmniNOCS creation is illustrated in Figure \ref{fig:dataset_prep} and explained below. A summary of the resulting OmniNOCS dataset is provided in Table \ref{table:nocs_datasets}.

\begin{table}[tb]
\centering
\begin{minipage}{.9\linewidth}
\resizebox{\linewidth}{!}{
\begin{tabular}{ l c c|c|c } 
 \hline
 3D Dataset                                 & Depth source & Instance mask source  & \#Images & \#Classes \\ 
 \hline
 KITTI \cite{Geiger2012CVPR}                & LiDAR & Human-labelled \cite{heylen2021monocinis} &  7.4k & 7 \\ 
 SUN-RGBD \cite{sunrgbd}                    & Depth Camera & {Segmentation model \cite{kirillov2023segany} \cellcolor{checkyes}} &  10k & 72 \\ 
 Objectron \cite{objectron2021}             & {NeRF \cellcolor{checkyes}} & {NeRF + Human label \cellcolor{checkyes}} &  132k & 9 \\ 
 nuScenes \cite{caesar2020nuscenes}         & LiDAR & {Segmentation Model \cite{cheng2020panoptic}\cellcolor{checkyes}} & 30k & 9 \\
 Hypersim \cite{hypersim}                   & Synthetic & Synthetic & 64k &  31 \\ 
 ARKitScenes \cite{dehghan2021arkitscenes}  & Depth Camera & {Segmentation model \cite{kirillov2023segany} \cellcolor{checkyes}} & 60k & 15 \\ \hdashline
 Cityscapes 3D \cite{cordts2016cityscapes}     & {Stereo\cellcolor{checkyes}} & Human-labelled & 3.4k & 8 \\ 
 Virtual KITTI \cite{cabon2020virtual}      & Synthetic & Synthetic & 4.1k & 3 \\
 NOCS-Real275 \cite{Wang_2019_CVPR}               & Depth camera &  Human-labelled &  7k & 6 \\ 
 Waymo OD \cite{waymodataset}               & LiDAR & Human-labelled &  62k &  7 \\ 
 \hline
 \textbf{OmniNOCS}                &  &  & \textbf{\numimages{}} & \textbf{\numclasses{}} \\
 \hline
\end{tabular}
}
\end{minipage}
\vspace{2mm}
\caption{\small{\textbf{Constituent 3D detection datasets used in OmniNOCS:} We augment many of these datasets with depth and masks, as they are missing in the original data (highlighted entries). In addition, we canonicalize the orientations of bounding boxes across all datasets.  \#Classes lists the number of classes we use from each dataset. Those above the dashed line are part of Omni3D\cite{brazil2023omni3d}.}}
\label{table:nocs_datasets}
\vspace{-8mm}
\end{table}

\noindent \textbf{Orientation canonicalization}: Although the datasets we use contain oriented 3D bounding boxes, they vary in their level of canonicalization. Some datasets have their class-canonical orientations i.e, all instances of a particular class are oriented consistently \emph{within the dataset} (for example, all cars in \cite{Geiger2012CVPR} have X axis pointing forward, and Z upwards). In such cases, we ensure that this canonicalization is consistent with all of OmniNOCS by applying a fixed class-specific offset orientation for the dataset. Datasets like Hypersim \cite{hypersim} have no class canonicalization at all, i\.e, although objects have tight bounding boxes, their XYZ axes directions are different for each instance. We manually label each object in such datasets to select the canonical orientation out of six possible orientations for the bounding box. More details on the labelling process are provided in the supplementary material. Some classes may have more than one choice for canonical orientations (due to symmetry), in which case an orientation is selected arbitrarily. 


\noindent \textbf{Depth augmentation}: For outdoor datasets, we use sparse depth from LiDAR if available. We recomputed the depth on Cityscapes using a state-of-the-art stereo depth model \cite{xu2023iterative}, as we found the depth from the original dataset to be noisy. Since  Objectron \cite{objectron2021} does not contain dense depth, we train Neural Radiance Fields \cite{mildenhall2020nerf, barron2022mipnerf360} for each video sequence in the dataset to obtain dense depth. 

\noindent \textbf{Instance segmentation}: Since many datasets we use were intended for 3D detection, they do not camera instance segmentations. We annotate ARKitScenes and SUNRGBD objects using instance masks from Segment-Anything (SAM) \cite{kirillov2023segany}, prompting it with the projected 3D bounding box. Although SUN-RGBD provides segmentation masks, we find that the SAM annotations are of superior quality. For Objectron, we create accurate masks by efficiently annotating in 3D space. Specifically, we apply the pipeline of~\cite{barron2022mipnerf360} to each of the Objectron videos. We fuse the resulting NeRF depth-maps to create a mesh. We post-process the 3D mesh, and get rid of the redundant vertices while keeping the vertices of the object. Finally, we create the masks for multiple frames by measuring the distance between each pixel on the depth map and the object mesh: pixels whose depth is far from the surface of the mesh are discarded, a process which also handles occlusions.

\noindent \textbf{NOCS computation}: As proposed in \cite{Wang_2019_CVPR}, the NOCS coordinate $x_{nocs}$ for a point on the object surface is defined as: 

\begin{gather}
    x_{nocs} = \frac{1}{s} \; ^{obj}T_{cam} \; x_{cam}
\end{gather}

\noindent where $^{obj}T_{cam} = [^{obj}{R}_{cam} | ^{obj}t_{cam} ]$ is the 6DoF transformation from the camera to the object frame, $x_{cam}$ the 3D location of the point in the camera frame, and $s$ is a scalar, the size of the diagonal of the object's tight bounding box. It can be interpreted as the object shape scaled to a box with a unit diagonal. For each image, starting from a 3D pointcloud ($x_{cam}$) obtained from the depth, we collect points on each object ($x^{i}_{cam}$) using its 3D bounding box and instance mask, which can be transformed to the NOCS coordinate ($x^{i}_{nocs}$) using (1). We store NOCS as a 3-channel 2D map (i.e $x^{i}_{nocs}$ at its corresponding 2D location obtained by projecting $x^{i}_{cam}$).  Our final result is the OmniNOCS dataset which contains instance segmentations, NOCS maps, and 3D bounding boxes for objects across \numclasses{} classes.

\section{\nocsmethod{} model}

We propose a novel architecture for monocular NOCS prediction termed ``\nocsmethod{}'' trained on OmniNOCS. \nocsmethod{} primarily uses self-attention layers, and is the largest trained monocular NOCS prediction model to date. \nocsmethod{} also contains a 3D size head and a learned PnP head that are used to estimate a canonically oriented 3D object bounding box from the NOCS.

\subsection{Model architecture}

\begin{figure*}[t]
    \centering
    \includegraphics[width=1\linewidth]{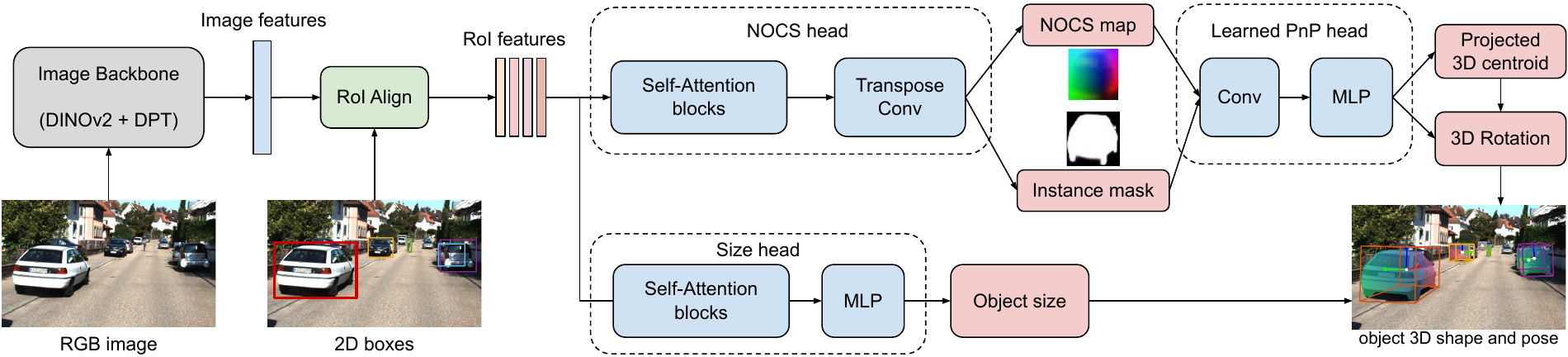}
    \vspace{-2mm}
    \caption
    {
    \small{
    \textbf{Architecture for \nocsmethod{}:} We use a transformer backbone to extract features from the input image, pool them using the 2D box RoIs, and feed the per-RoI features to the NOCS and size heads. Our novel NOCS head jointly predicts the NOCS and instance mask for the RoI. Our learned PnP head for pose estimation uses the predicted NOCS and instance mask to predict the projected 3D coordinate and 3D rotation of the object.
    }
    }
\label{fig:model_arch}
    \vspace{-4mm}
\end{figure*}

As shown in Figure \ref{fig:model_arch}, the \nocsmethod{} architecture comprises an image backbone, a NOCS head, a size head, and a learned PnP head. 

\noindent \textbf{Backbone:} Our backbone is a frozen Vision Transformer (ViT) \cite{dosovitskiy2020image}, that uses DINOv2 weights \cite{oquab2023dinov2} from self-supervised pretraining. This choice is motivated by recent findings on using frozen DINOv2 backbones for understanding  depth, multi-view correspondences, or relative pose  \cite{goodwin2022zero, zhang2023tale}. Additionally, we use 
a Dense Prediction Transformer (DPT) \cite{ranftl2021vision} architecture to fuse low-resolution DINOv2 representations from multiple intermediate layers and upsample them by a factor of 8. This higher feature resolution is desirable to improve predictions on smaller objects. We train the DPT layers while keeping the ViT layers frozen. We use 2D input boxes to sample the DPT features using RoIAlign ~\cite{he2017mask} into a 28x28 fixed resolution grid that is fed to the NOCS and size heads. While DPT has been previously leveraged for tasks that provide dense supervision, such as semantic segmentation and depth prediction, we use it in an instance-level prediction context, training it using supervision from object RoIs. 

\noindent \textbf{NOCS head:} All existing NOCS prediction models \cite{Wang_2019_CVPR, li2019cdpn, wang2021gdr} use MaskRCNN heads \cite{he2017mask} with convolutional layers, in some cases with a separate head for each NOCS coordinate \cite{Wang_2019_CVPR}. While this works on small datasets, we find that it significantly limits performance on OmniNOCS. \nocsmethod{} therefore uses a simple (but large) transformer NOCS head with 10 self-attention blocks and one final upsampling conv layer to jointly predict both the instance mask and the NOCS map in the same output head. Similar to \cite{Wang_2019_CVPR}, for each NOCS coordinate, we predict logits over non-overlapping bins, and use the softmax to obtain the final NOCS value. Our ablations show that this choice works best on our challenging OmniNOCS dataset.

\noindent \textbf{Size and learned PnP heads:} We use a size head to predict the 3D size to scale the NOCS predictions to metric object coordinates. Our size head also uses self-attention blocks. Contrary to previous work \cite{Wang_2019_CVPR, 3DRCNN_CVPR2018, brazil2023omni3d}, it does not use any class-specific layers or per-class average size statistics, as these limit scalability to larger and diverse vocabularies. Our learned PnP head is inspired from \cite{li2019cdpn, wang2021gdr} and uses convolutional layers to predict the 3D orientation and a 2D projection of the 3D centroid. The orientation is predicted using the 6D partial rotation matrix representation \cite{zhou2019continuity} in allocentric space \cite{3DRCNN_CVPR2018, brazil2023omni3d}. 

\subsection{Localization from NOCS predictions}
\label{subsec:localization}

To localize the object in 3D, we first scale the predicted NOCS using the predicted size to obtain unnormalized metric object coordinates. \nocsmethod{}'s learned PnP head predicts the object orientation in allocentric space, which is converted to the egocentric frame using the predicted projection of the 3D centroid. Using the unnormalized 3D object coordinates, object orientation, and the 3D centroid ray, the object depth is estimated from the corresponding 2D points. 

Alternatively, the 6DoF pose for the object can be solved directly (without using the learned PnP head) by solving a PnP problem using efficient solvers \cite{Lepetit:160138}. However, the orientations so obtained are known to be more sensitive to errors in NOCS coordinates, and less robust when compared to learned methods \cite{li2019cdpn, wang2021gdr}. 

\subsection{Losses}
\label{subsec:nocs-losses}

When training \nocsmethod{}, we supervise the NOCS, mask, 3D size, orientation and centroid predictions using appropriate losses. We supervise predicted masks using a simple L2 loss with respect to the ground truth instance mask. Since our NOCS head uses binned prediction where the final NOCS value is a softmax over non-overlapping bin logits, we use a combination of cross-entropy and regression losses. 

\vspace{-1em}
\begin{gather*}
    \mathcal{L}_{mask} = || \texttt{mask}_{pred} - \texttt{mask}_{gt} ||_2 \\
    \mathcal{L}_{NOCS} = \texttt{softmaxCE}  (\hat{\mathbf{n}}, \hat{\mathbf{n}}_{gt}) + ||\mathbf{n} - \mathbf{n}_{gt} ||_1
\end{gather*}

\noindent where $\hat{\mathbf{n}}$ are the predicted logits over the discretized bins and $\mathbf{n}$ the continuous NOCS coordinate prediction. 

In addition to the supervised loss above, we also use a variant of the self-supervised NOCS loss $\mathcal{L}_{ss}$ from \cite{monorun2021} that minimizes the reprojection error of the predicted NOCS using the ground truth pose and predicted mask. More details about $\mathcal{L}_{ss}$ are in the supplementary material. The NOCS and mask losses are applied to the fixed resolution grid predictions. Since our NOCS ground truth can be sparse, it is applied at only those locations on the grid that have valid NOCS ground truth. 

Since our 3D size head also uses binned prediction, we use a combination of softmax cross-entropy and L1 loss for supervision. Note that our size loss is normalized by the ground truth size, in order to penalize errors on smaller objects equally. For the learned PnP head, we also supervise the centroid and rotation predictions using L1 losses in their output space. 

\vspace{-1em}
\begin{gather*}
    \mathcal{L}_{size} = \texttt{softmaxCE}  (\hat{\mathbf{s}}, \hat{\mathbf{s}}_{gt}) + |\mathbf{s} - \mathbf{s_{gt}}| / \mathbf{s_{gt}} \\    
    \mathcal{L}_{rot} = \| {^c\mathbf{R}}_{o_{pred}} - {^c\mathbf{R}}_{o_{gt}} \|_
    {1,1}\\
    \mathcal{L}_{centroid} = || \mathbf{c}_o - \mathbf{c}_{o_{gt}} ||_2 \\
    \mathcal{L}_{PnP} = \mathcal{L}_{rot} + \mathcal{L}_{centroid}
\end{gather*}

The total loss for training \nocsmethod{} is a weighted sum: 
\begin{gather*}
    \mathcal{L}_{total} = w_{size} \mathcal{L}_{size} +  w_{mask} \mathcal{L}_{mask} +  w_{NOCS}  \mathcal{L}_{NOCS} + w_{ss} \mathcal{L}_{ss} + w_{PnP} \mathcal{L}_{PnP}
\end{gather*}

\section{Experiments}
\label{sec:experiments}

Although previous works predict NOCS accurately on a few categories \cite{Wang_2019_CVPR, li2019cdpn, wang2021gdr}, they only evaluate on  3D detection (localization) or pose estimation tasks, without quantifying the accuracy of the predicted NOCS. In section~\ref{sec:exp-nocs-prediction}, we propose metrics and establish a benchmark to evaluate NOCS on the OmniNOCS dataset. We  compare the localization accuracy of \nocsmethod{} against existing benchmarks, by evaluating its localization accuracy on nuScenes \cite{caesar2020nuscenes} in section~\ref{exp:localization}. We also evaluate the unique ability of \nocsmethod{} to transfer to unseen datasets and domains. Finally, we provide ablations on critical design choices made in our model in section~\ref{sec:results-ablations}. 

\begin{figure}[t]
    \includegraphics[width=1\linewidth]{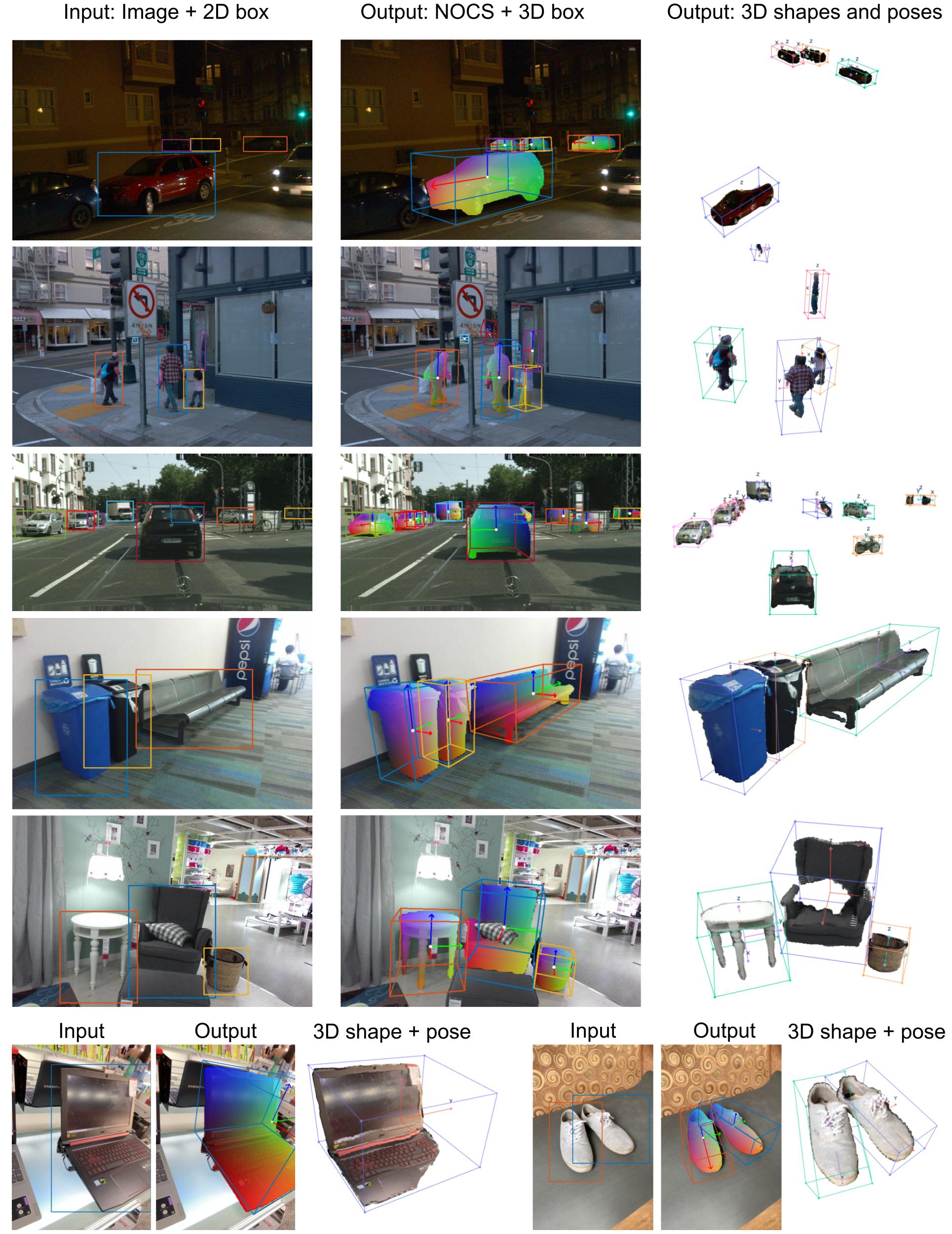}
    \vspace{-2mm}
    \caption{\small{
    \textbf{Example results of our single unified \nocsmethod{} model across various datasets and object classes}.
    The left column shows input images and query 2D bounding boxes. The center column shows the NOCS+instance maps predicted by \nocsmethod{} along with the estimated 3D pose overlaid on the input image. The NOCS can be used with the 3D boxes and object size to lift the objects to a 3D pointcloud, which is shown in the right column. Last row contains two examples.
    }
    }
\label{fig:qual_nocs_1}
\end{figure}
\FloatBarrier

\textbf{Datasets:} For our experiments, we train \nocsmethod{} on the OmniNOCS dataset containing \numclasses{} classes, holding out images from the NOCS-Real275 dataset. We use NOCS-Real275 to evaluate \nocsmethod{}'s cross-dataset generalization capabilities by not training on it. NOCS-Real275 has 6 classes that overlap with the rest of OmniNOCS, although they are not the top 20 classes. It also differs from the rest of OmniNOCS in terms of camera parameters and the context of objects in the scene.

\subsection{NOCS prediction accuracy}
\label{sec:exp-nocs-prediction}

We propose the use of NOCS mAE and NOCS mPSNR to evaluate the accuracy of predicted NOCS. For each object the mean Absolute Error (mAE) and mPSNR are computed for all points in the intersection of ground truth and predicted masks, and reported as a mean of per-category means. However, the NOCS mAE/mPSNR do not penalize undersegmentations / sparse predictions. Since we also would like the NOCS predictions to span the full visible instance, we also evaluate the 2D mask mIoU.

We evaluate \nocsmethod{} on a subset of OmniNOCS that contains 75 classes with accurate ground truth NOCS and masks, see Table \ref{tab:exp-nocs}. It is able to predict NOCS with errors less than 9\% (0.089 mAE on OmniNOCS). We also evaluate on the held-out NOCS-Real275 dataset, which was not used to train \nocsmethod{}. We find that the zero-shot \nocsmethod{} outperforms the NOCS baseline from \cite{Wang_2019_CVPR} that was fully trained on this dataset, a strong indication of the generalization capabilities of \nocsmethod{}.

\begin{table}[tb]
\centering
\setlength{\tabcolsep}{3pt}
\def\arraystretch{1}
\resizebox{\linewidth}{!}{
\begin{tabular}{l|c c c|c c c}
\hline
  &  \multicolumn{3}{c|}{NOCS-Real275} & \multicolumn{3}{c}{OmniNOCS}  \\ 
 Method &  NOCS MAE$\downarrow$ & NOCS PSNR$\uparrow$  & Mask IoU$\uparrow$  & NOCS MAE$\downarrow$  & NOCS PSNR$\uparrow$  & Mask IoU$\uparrow$   \\ 
 \hline
 NOCS baseline \cite{Wang_2019_CVPR} &  0.121  & 16.345  & 86.10 & - & - & - \\
\textbf{\nocsmethod{}} & \textbf{0.107} & \textbf{18.527} & \textbf{89.03} & \textbf{0.094} & \textbf{20.245} & \textbf{78.50} \\
 \hline
\end{tabular}
}
\vspace{3mm}
\caption{\small{\textbf{NOCS and mask prediction evaluation}: We report metrics for \nocsmethod{} on 75 classes of  OmniNOCS: it is the first model that is capable on predicting NOCS on such diverse data. Additonally, \nocsmethod{} outperforms \cite{Wang_2019_CVPR} which trained on NOCS-Real275, without training on NOCS-Real275.}}
\label{tab:exp-nocs}
\vspace{-4mm}
\end{table}




\subsection{3D localization accuracy}
\label{exp:localization}

NOCS provides dense 3D-2D correspondences that can be used to estimate the 3D object oriented bounding box. This is done by using the predicted object size and solving a PnP problem, as explained in Sec.~\ref{subsec:localization}.  We evaluate the accuracy of our estimated bounding boxes in different settings:

\textbf{Outdoor scenes:} We compare \nocsmethod{}'s 3D localization accuracy to that of other 3D detection models using the nuScenes true positive localization metrics on the challenging nuScenes dataset (included in OmniNOCS). We compare to \cubercnn{}, since it is the only other model that generalizes across diverse datasets. However, \cubercnn{} has two notable differences to \nocsmethod{}: 1) being a detection model, it also jointly detects 2D object regions,  and 2) it uses Chamfer loss causing it to have high orientation errors. As a more comparable baseline, we use a variant of our model (termed  "\boxmethod{}"), by replacing our NOCS head with the cube head from \cubercnn{} without a Chamfer loss. It uses the input 2D boxes. More details about \boxmethod{} are in the supplementary material. From Table \ref{table:outdoor-localize}, we find that \nocsmethod{}'s box orientations are canonical and more accurate than both \cubercnn{} and \boxmethod{}. Moreover, the boxes estimated from \nocsmethod{}'s NOCS predictions are comparable in translation and scale errors to those of \cubercnn{} and \boxmethod{}, even though it does not directly regress depth. While all 3 methods generalize across several classes and domains, baselines trained solely on nuScenes localization (top half) significantly outperform them on this task.


\begin{table}[tb]
\centering
\begin{minipage}{.9\linewidth}
\setlength{\tabcolsep}{3pt}
\def\arraystretch{1}
\resizebox{\linewidth}{!}{
\begin{tabular}{l c|c c c c}
 \hline
 Method & Multi-dataset & mATE (m)$\downarrow$ & mAOE (rad)$\downarrow$ & mASE (\%)$\downarrow$ & mIoU$\uparrow$ \\ [0.5ex] 
 \hline
 FCOS3D \cite{wang2021fcos3d} & \xmark & 0.777 & 0.400 & \textbf{0.231} & -\\
 PGD \cite{wang2022probabilistic} & \xmark & 0.675 & 0.399 & 0.236 & -\\ 
 EProPNP \cite{chen2022epro} & \xmark & 0.676 & 0.363 & 0.263 & -\\
 EProPNP + TTA \cite{chen2022epro} & \xmark & \textbf{0.653} & \textbf{0.319} & 0.255 & -\\ [0.5ex]
  \hline
 \cubercnn{} \cite{brazil2023omni3d} & \cmark &  \textbf{0.650} &  1.305 & \textbf{0.283} & 0.349 \\
 \boxmethod{} & \cmark & 0.790 & 0.573 & 0.301 & 0.280 \\
 \nocsmethod{} & \cmark &  0.887 & \textbf{0.558} & 0.291 & \textbf{0.377} \\ [1ex]
 \hline
\end{tabular}
}
\end{minipage}
\vspace{3mm}
\caption{\footnotesize{\textbf{Outdoor localization:} Comparison of 3D localization accuracy on the nuScenes subset of OmniNOCS. Top half shows methods that are only trained on nuScenes: while these perform better on nuScenes itself, they do not generalize to other datasets and classes. OmniNOCS is competitive with \cubercnn{} on mIoU while also predicting canonical orientations and object coordinates. Note that mIoU and orientation error only uses the predicted yaw orientation component.}}
\label{table:outdoor-localize}
\vspace{-4mm}
\end{table}

\textbf{Cross-dataset generalization (indoor):} Here, we hold out NOCS-Real275 from OmniNOCS when training \nocsmethod{}, and use it evaluate \nocsmethod{}'s ability to generalize to unseen domains. NOCS-Real275 features a tabletop setting with multiple objects, which is semantically different from our other indoor datasets. We compare against \cubercnn{} \cite{brazil2023omni3d} and the NOCS \cite{Wang_2019_CVPR} model that was trained from scratch on this dataset alone. \cite{Wang_2019_CVPR} also uses class-specific model parameters and additional losses for symmetric objects that do not scale to larger vocabularies.

The results are shown in Table \ref{table:zero-shot-localize}. We find that \nocsmethod{} is more accurate at transferring to this unseen dataset compared to \cubercnn{}, indicating the generalizability of NOCS-based localization methods over that of box-regression methods. Note that the mAP metrics used in the NOCS-Real275 datasets may also be affected by the false positives/negatives from \cubercnn{}. However, both models are worse that the supervised NOCS baseline.

\begin{figure}[t]
    \centering
    \includegraphics[width=1\linewidth]{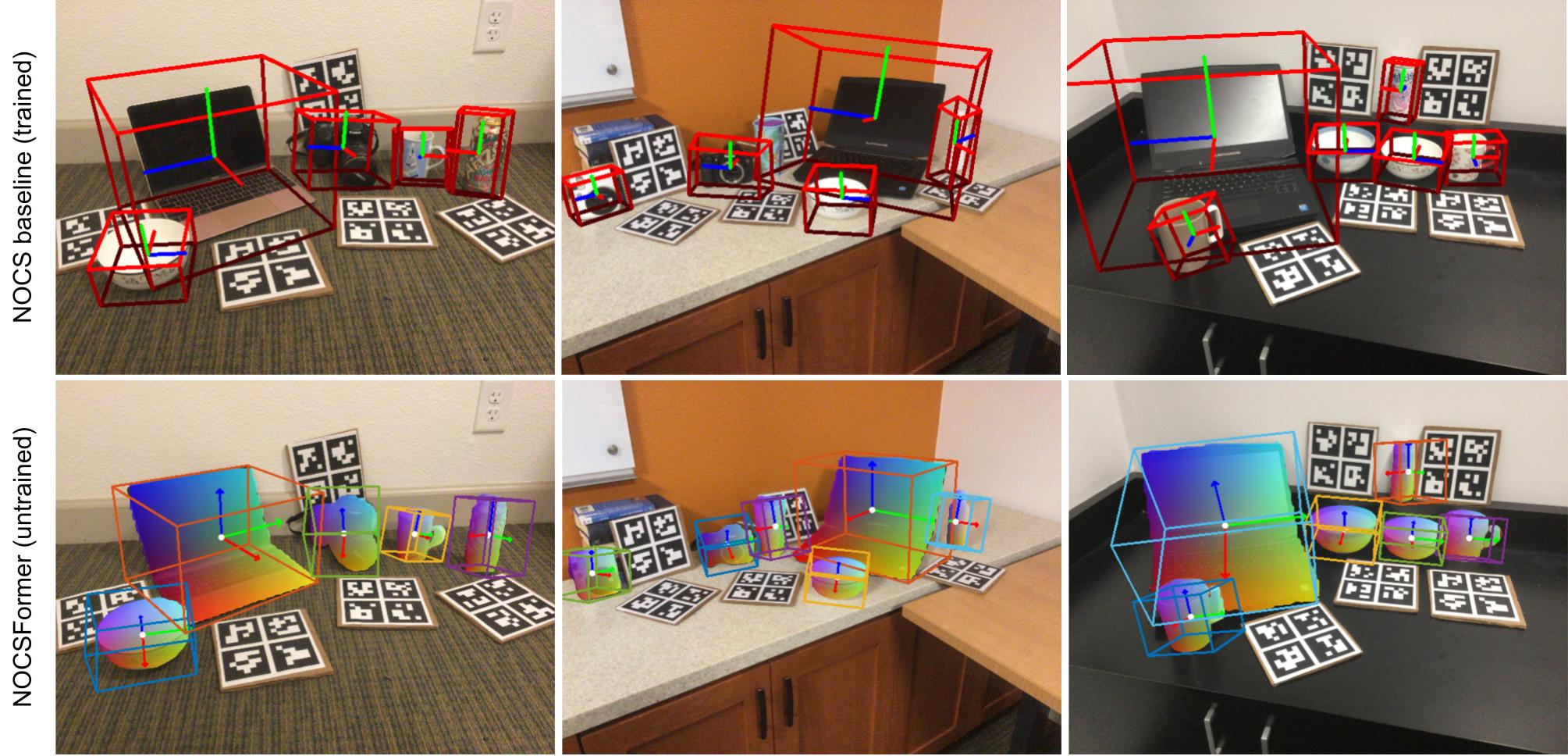}
    \vspace{-3mm}
    \caption
    {
    \small{
    \textbf{Generalization across datasets}: \nocsmethod{} can generalize to new datasets that present new camera models and object domains. We show this zero-shot capability of \nocsmethod{} (bottom row) on the NOCS-Real275 test set without training on NOCS-Real275 dataset. The predictions from a NOCS baseline~\cite{Wang_2019_CVPR} trained explicitly on the NOCS-Real275 dataset are shown in the top row.
    }
    }
\label{fig:nocs-zero-shot}
\end{figure}


\begin{table}[b]
\centering
\begin{minipage}{.65\linewidth}

\setlength{\tabcolsep}{3pt}
\def\arraystretch{1}
\resizebox{\linewidth}{!}{
\begin{tabular}{l c c|c c c} 
 \hline
  &  & &  \multicolumn{2}{c}{mAP}  \\ [0.5ex] 
 Method & Transferred? & Depth input? & 3D IoU @ 25 & 3D IoU @ 50  \\ [0.5ex] 
 \hline
 NOCS model \cite{Wang_2019_CVPR} & \xmark & \redcmark & \textbf{79.6} & \textbf{72.4} \\ [0.5ex] 
 \hline 
 CubeRCNN \cite{brazil2023omni3d} & \cmark & \greenxmark & 14.9 & 4.1 \\
 \nocsmethod{} & \cmark & \greenxmark & \textbf{43.5} & \textbf{10.6} \\
 \hline
\end{tabular}
}
\end{minipage}
\vspace{2mm}
\caption{\small{\textbf{Cross-dataset generalization}: Comparison of localization accuracy on NOCS-Real275 for all classes, using mAP at different thresholds. This dataset has been held out when training CubeRCNN and \nocsmethod{}. \nocsmethod{} is able to generalize to NOCS-Real275 \emph{without any additional training}.}} 
\label{table:zero-shot-localize}
\vspace{-4mm}
\end{table}

\subsection{3D orientation accuracy}
\label{exp-orientation}

A key challenge for 3D localization models trained on multiple datasets \cite{brazil2023omni3d} is predicting canonical object orientations, as these datasets typically differ in their object pose conventions. OmniNOCS ensures that object poses in all constituent datasets are consistent within a category, enabling our model to predict canonical orientations. We evaluate this by computing the accuracy at different orientation error threshold on KITTI-val \cite{chen20173dop} in Table \ref{tab:exp-orientation}. We compare the accuracy of our orientation predictions to those of \cubercnn{} \cite{brazil2023omni3d}, which are trained using Chamfer loss, and therefore suffers at predicting canonical orientations. We also compare against our \boxmethod{} baseline that directly supervises the orientations. We find that orientations estimated using \nocsmethod{} are more consistent and accurate than both our baselines. 

\begin{table}[b]
\centering
\begin{minipage}{.9\linewidth}
\setlength{\tabcolsep}{3pt}
\def\arraystretch{1}
\resizebox{\linewidth}{!}{
\begin{tabular}{l|c c c|c c c} 
 \hline
        & \multicolumn{3}{c|}{Gravity axis} & \multicolumn{3}{c}{Heading axis (X)} \\
         & Accuracy & Accuracy & Accuracy & Accuracy & Accuracy & Accuracy \\ 
 Method &  @ 1 deg & @ 5 deg & @ 90 deg & 5 deg & @ 10 deg & @ 90 deg \\ 
 \hline
 CubeRCNN   &   5.75    & 18.29 & 51.74 & 19.43 & 21.55 & 34.17 \\ 
 \boxmethod{} & 80.79 & 98.52 & 100.0 & 41.99  & 55.88 & 77.26 \\
\textbf{\nocsmethod{}} &  \textbf{84.24}   & \textbf{99.95} &  \textbf{100.0} & \textbf{49.16} & \textbf{59.65} & \textbf{81.99} \\
 \hline
\end{tabular}
}
\end{minipage}
\vspace{3mm}
\caption{\small{\textbf{3D orientation accuracy for models trained on multiple datasets:} Models that use Chamfer distance (CubeRCNN) for supervising orientation heads end up being inconsistent in their orientation predictions. The results are averaged over 5 classes in the KITTI-val subset \cite{chen20173dop}.}}
\label{tab:exp-orientation}
\end{table}

\subsection{Ablations}
\label{sec:results-ablations}

\begin{table}[t]
\centering
\begin{minipage}{.6\linewidth}
\resizebox{\linewidth}{!}{
\setlength{\tabcolsep}{3pt}
\def\arraystretch{1}
\begin{tabular}{l|c c c}
\hline
 Architecture  & NOCS PSNR$\uparrow$ & Mask IoU$\uparrow$  \\ 
\hline
 \textbf{\nocsmethod{}}    & \textbf{20.431} & \textbf{81.69} \\ 
 Hourglass head    &  -3.064  &   -3.16   \\
 MaskRCNN head     &  -5.23   &    -21.27 \\
 w/o discretized prediction  &   -4.11    &    -   &        \\ 
\hline
\end{tabular}
}
\end{minipage}
\vspace{6mm}
\caption{\small{Comparison of different architecture choices for the NOCS and mask prediction head: We experiment with transformer, hourglass and MaskRCNN architectures. We compare whether discretized prediction of NOCS is better than continuous regression for transformer heads. We quantify the difference in performance when using separate heads for mask and NOCS prediction as opposed to a single head.}}
\label{tab:exp-sizerot}
\end{table}


\textbf{NOCS head architecture}: While all existing architectures use a few CNN layers for their NOCS heads \cite{Wang_2019_CVPR}, we find that this significantly limits the NOCS prediction performance when scaling to more classes. Using our transformer NOCS head provides a 5.23dB improvement in NOCS PSNR and a 21.27\% improvement in mask mIoU. We also experiment with a larger convolutional head: the Hourglass model from \cite{birodkar2021surprising}. We find that the \nocsmethod{} NOCS heads are even better than Hourglass, with an improvement of 3.06dB on NOCS PSNR and 3.16\% on mask IoU (as shown in Table \ref{tab:exp-sizerot}). 

\noindent \textbf{Discretized versus continuous prediction of NOCS}: \nocsmethod{} models the final NOCS prediction as a classification layer with 50 non-overlapping bins. This was observed to be better than a linear regression layer with a MaskRCNN head in \cite{Wang_2019_CVPR}. We find that it is also significantly better when using a transformer head, improving the NOCS prediction PSNR by 4.11 dB, as in Table \ref{tab:exp-sizerot}.

\section{Conclusion and Future work}

This paper has introduced a new large scale dataset of Normalized Object Coordinates (NOCS) for a wide variety of object classes in indoor and outdoor scenes. It has also proposed a single transformer-based model \nocsmethod{} that can predict object 3D pose, size, and NOCS for any of these objects given 2D bounding box inputs. These allows our model to obtain 3D shape and oriented bounding boxes of objects in metric scale from a single input image. This represents the first attempt to generalize NOCS estimation beyond small datasets of narrow domain, increasing the number of object categories available by an order of magnitude. We hope this provides a means for others to explore large-scale monocular 3D object pose and shape estimation. 

Some limitations of our current method include the handling of symmetric objects, where the coordinate system has multiple possible solutions, and reflected geometry, such as left / right shoes. Future work could address these issues, for example, minimizing over multiple coordinate frames in the loss for symmetric objects, and potential estimation of left / right coordinate frames for reflected geometry.  

 

\pagebreak
\appendix
\begin{center}
\large \textbf{Appendix}
\end{center}

This appendix provides additional qualitative results, including results on in-the-wild internet image collections. Detailed OmniNOCS statistics and more information about the annotation process are provided in Section B. Section C contains more details on NOCSformer architecture and training. 

\vspace{-2mm}
\begin{figure}[h]
\centering
    \includegraphics[width=0.95\linewidth]{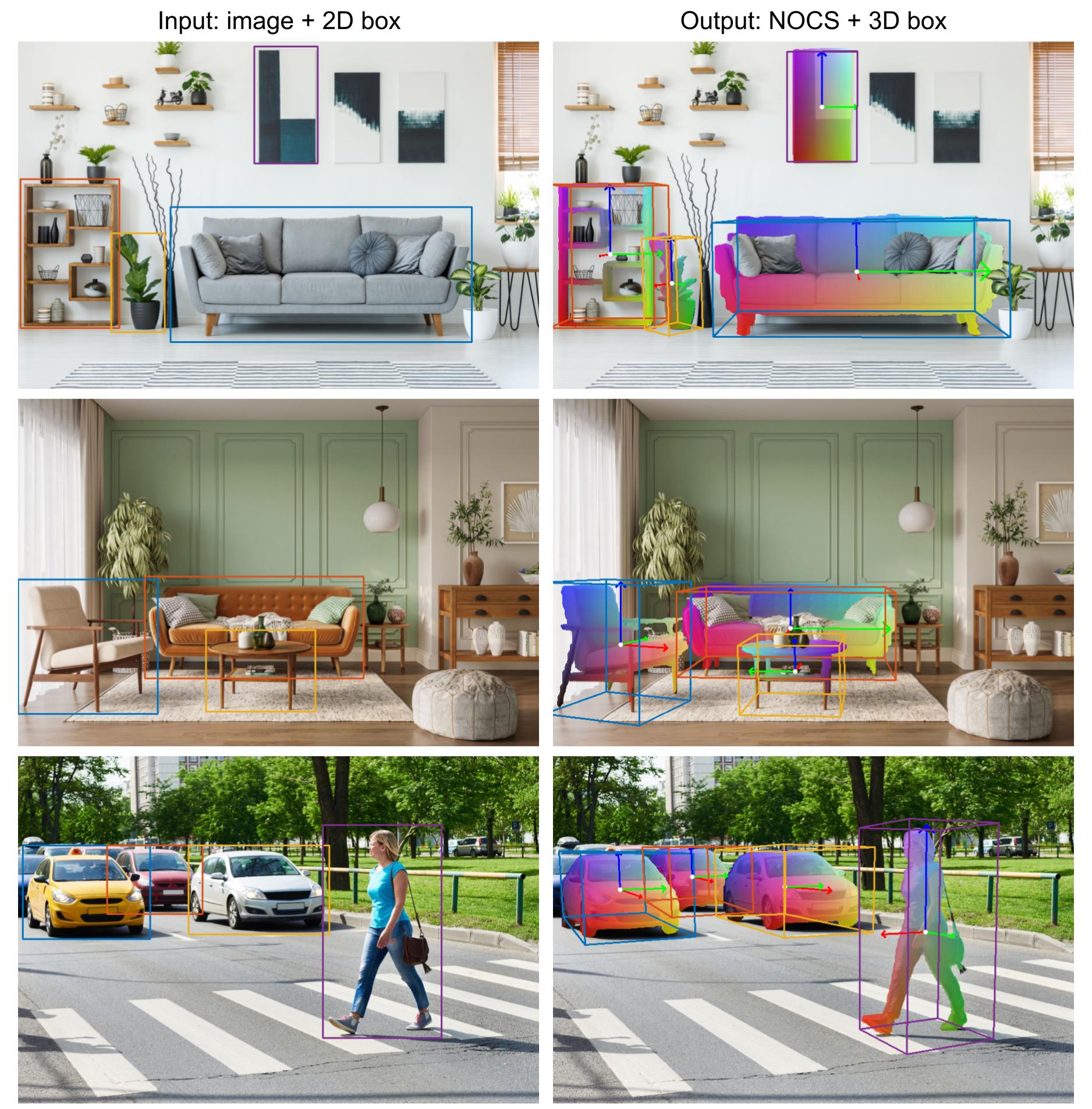}
    \captionof{figure}
    {
    \small{
    NOCSformer trained on OmniNOCS dataset generalizes to web \textbf{stock images} that are outside the training dataset. The predicted NOCS and 3D bounding boxes are shown on right for each input image and 2D query. These results demonstrate the capability of NOCSformer trained on OmniNOCS for 3D object reconstruction of in-the-wild images.
    }
    }\label{fig:suppl_web_images}
\end{figure}%
\vspace{-4mm}

\section{Qualitative results of {\nocsmethod}}

\begin{figure}[t]
    \includegraphics[width=\linewidth]{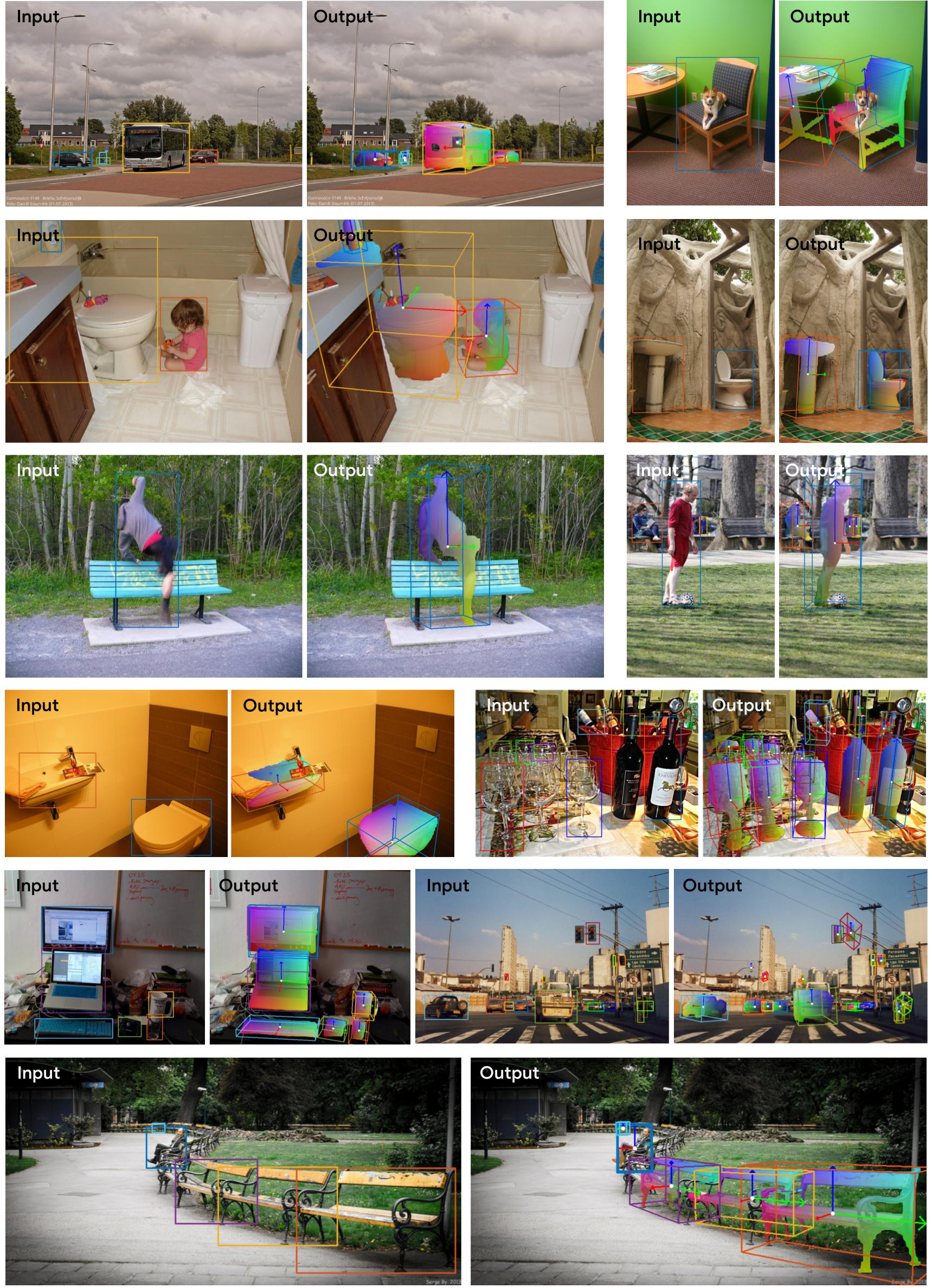}
    \captionof{figure}
    {
    \small{
    Results of \nocsmethod{} (trained on OmniNOCS dataset) on images from the \textbf{COCO dataset}. For each image pair, the left image shows the input 2D bounding box, and the right image shows the predicted NOCS and 3D bounding box. Note that COCO is \textbf{not} part of OmniNOCS. However, the model generalizes to the  challenging COCO images, predicting NOCS and 3D bounding boxes from 2D queries. The model is also able to generalize to unseen (though related) classes -- for example generalizing to park benches when only trained on sofas and chairs.
    }
    }\label{fig:suppl_coco_images}
\end{figure}%

\begin{figure}[t]
    \includegraphics[width=\linewidth]{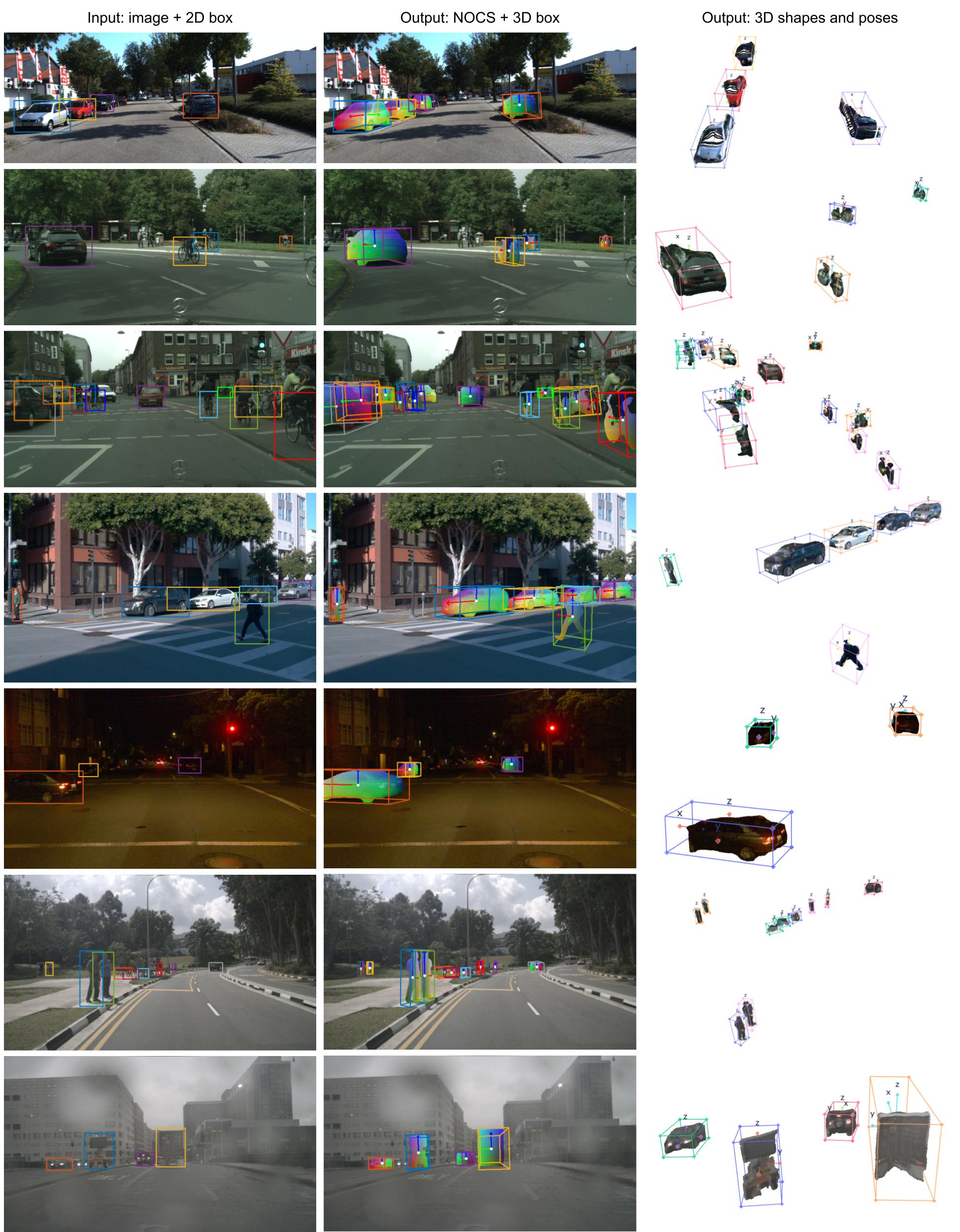}
    \captionof{figure}
    {
    \small{
    Qualitative results of \textbf{\nocsmethod} model for various outdoor datasets. Left column shows the input images and query 2D bounding boxes. The center image shows the predicted per object NOCS, segmentation, and 3D oriented bounding box from our model corresponding to each input query overlaid on the input image. The right image shows the object coordinates lifted to 3D using the predicted 6DoF pose.
    }
    }\label{fig:supplm_qual_outdoors}
\end{figure}

\begin{figure}[t]
    \includegraphics[width=\linewidth]{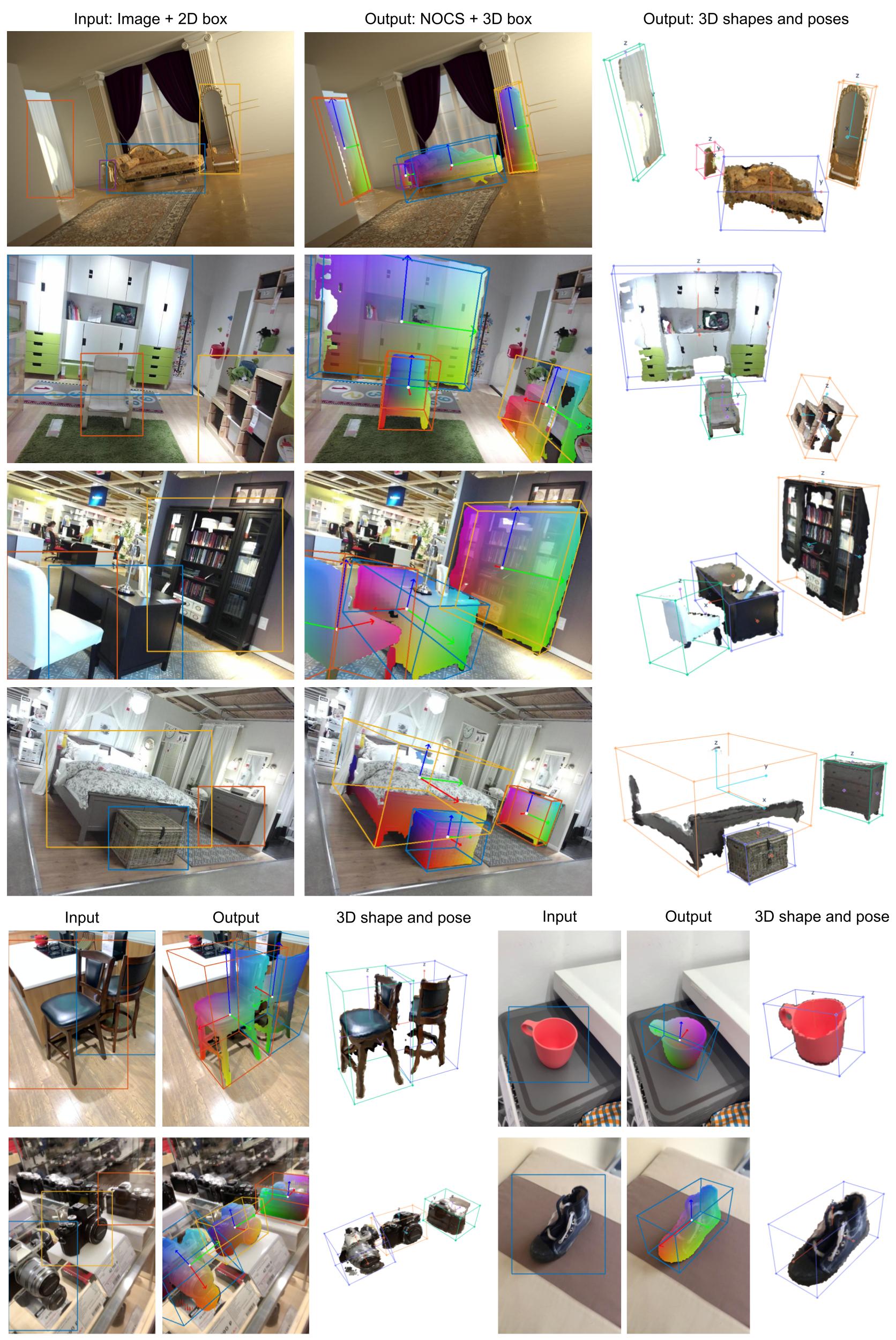}
    \captionof{figure}
    {
    \small{
    Qualitative results of \textbf{\nocsmethod} model for various indoor datasets. Left column shows the input images and query 2D bounding boxes. The center image shows the predicted per object NOCS, segmentation, and 3D oriented bounding box from our model corresponding to each input query overlaid on the input image. The right image shows the object coordinates lifted to 3D using the predicted 6DoF pose.
    }
    }\label{fig:suppl_qual_indoors}
\end{figure}

We present additional qualitative results for {\nocsmethod} trained in different settings -- outdoor scenes (Fig.  \ref{fig:supplm_qual_outdoors}) and indoor scenes, (Fig. \ref{fig:suppl_qual_indoors}). We also provide results on in-the-wild images from the web (Fig. \ref{fig:suppl_web_images}) and some images from the COCO dataset (Fig. \ref{fig:suppl_coco_images}). 


\FloatBarrier

\subsection{Generalization to in-the-wild images}

Fig. \ref{fig:suppl_web_images} shows NOCS predictions on online stock images, highlighting the model's ability to generalize beyond the training contexts. Additionally, we run our model on the COCO dataset \cite{cocodataset}, and show results in Fig. \ref{fig:suppl_coco_images}. 
The results confirm these findings on the model's generalization capabilities. 
Since our model does not contain any class specific parameters, it can also be queried on classes it is not trained on. In such cases, it tends to perform reliably for classes that are closely related to the training classes. For example, it performs well on park benches in Fig. \ref{fig:suppl_coco_images} despite only being trained on couches, sofas and chairs. 

\subsection{Additional results on OmniNOCS}

Fig. \ref{fig:supplm_qual_outdoors} shows results from KITTI \cite{Geiger2012CVPR}, Cityscapes \cite{cordts2016cityscapes}, Waymo \cite{waymodataset}, and nuScenes \cite{caesar2020nuscenes} for different object classes including cars, bikes, pedestrians, and trucks. Our method is able to predict NOCS and 3D bounding boxes reliably even under occlusions, during the night time, or in severe rainy conditions. We use the NOCS and 3D bounding box to lift the object to a 3D pointcloud. The rightmost 3D column shows that the obtained pointcloud respects relative 3D distances and orientations. 
Fig. \ref{fig:suppl_qual_indoors} shows results from Objectron \cite{objectron2021}, Hypersim \cite{hypersim} and SUN-RGBD \cite{sunrgbd} in indoor environments, taken from different types of cameras. This also includes results on some less frequent classes such as bookshelves, curtains, and mirrors.

\section{OmniNOCS dataset details}

\subsection{Data statistics}

\begin{figure}[h]
  \centering
  \includegraphics[width=\linewidth]{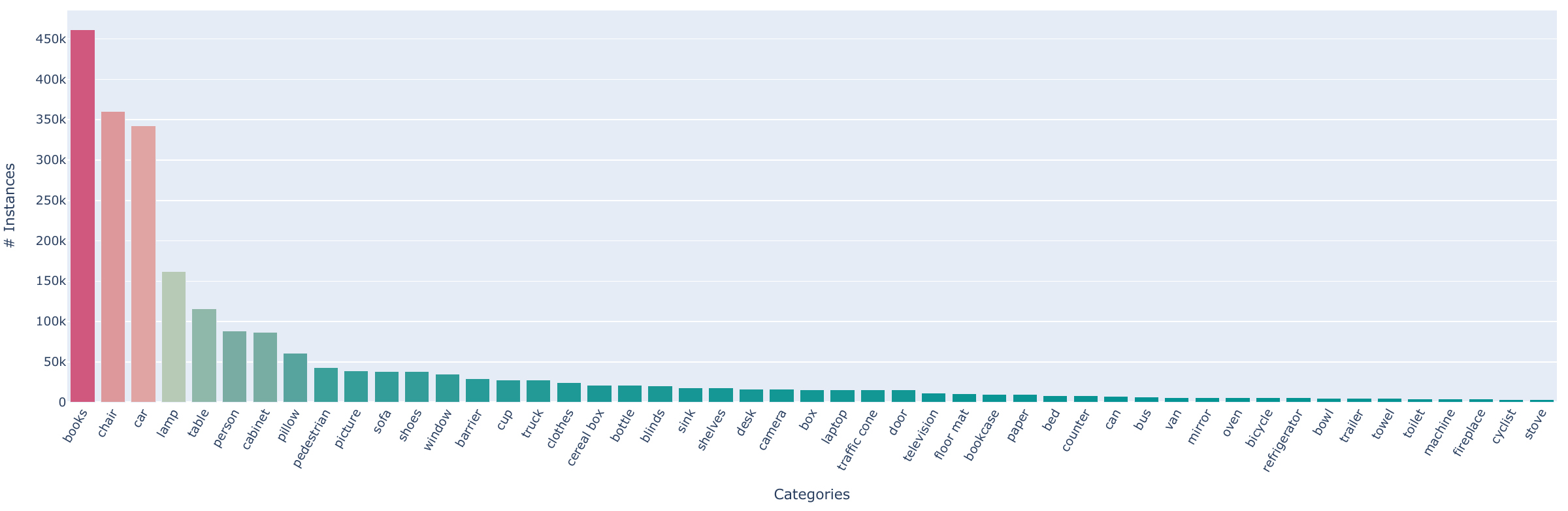}
  \includegraphics[width=\linewidth]{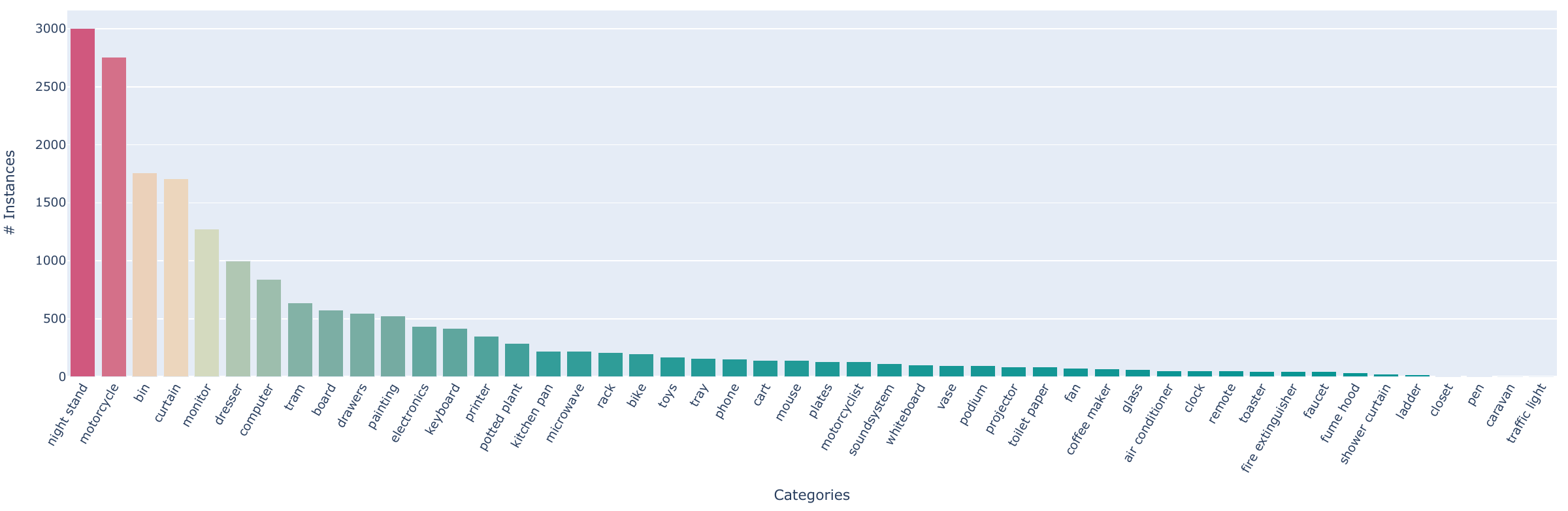}
  \vspace{-2em}
    \caption{\small{A histogram of the number of instances for each category in OmniNOCS: (top) most frequent categories (bottom) least frequent categories.}}
   \label{fig:instance-histogram}
\end{figure}

\begin{figure}
\centering
\begin{subfigure}{0.32\textwidth}
    \includegraphics[width=\textwidth]{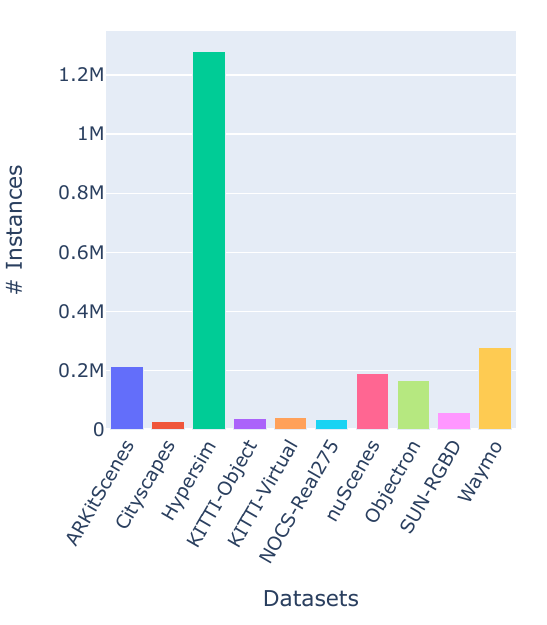}
    \label{fig:first}
\end{subfigure}
\hfill
\begin{subfigure}{0.32\textwidth}
    \includegraphics[width=\textwidth]{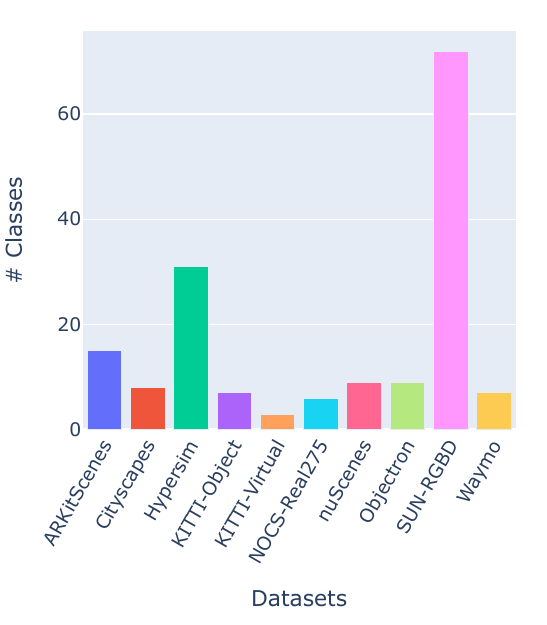}
    \label{fig:second}
\end{subfigure}
\hfill
\begin{subfigure}{0.32\textwidth}
    \includegraphics[width=\textwidth]{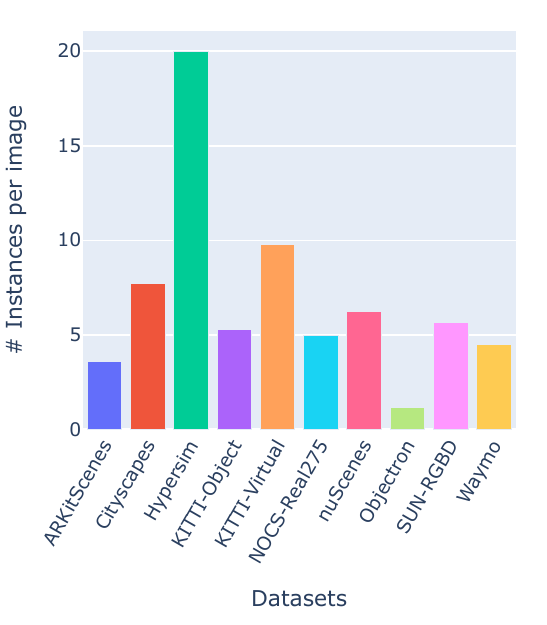}
    \label{fig:third}
\end{subfigure}
\vspace{-5mm}
\caption{\small{Statistics for the datasets that are used in OmniNOCS. These datasets span different domains and classes. They also also differ significantly in the number of instances per image.}}
\label{fig:data-stats}
\end{figure}

OmniNOCS has more than 2.2M object instances spanning the train, val and test splits (not counting repeated instances). 
A histogram of the number of instances in the top and bottom 50 categories is shown in Fig. \ref{fig:instance-histogram}. 

\begin{figure}[t]
  \centering
  \includegraphics[width=\linewidth]{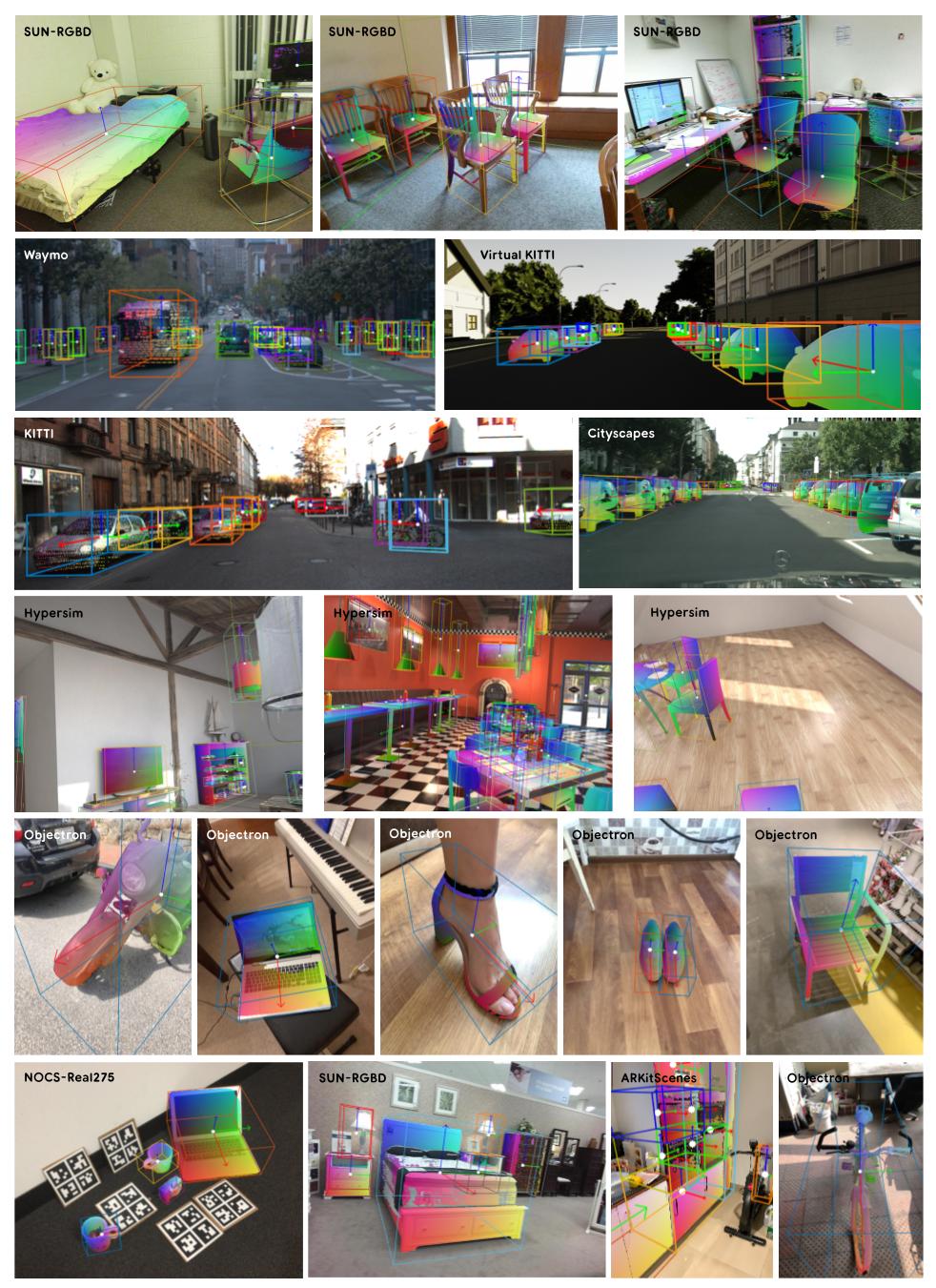}
   \caption{\small{\textbf{Example annotations from our OmniNOCS dataset}: Each frame contains multiple objects, with each object having a 3D bounding box with canonical orientations, NOCS coordinates and  instance mask.}}
   \label{fig:dataset-fig}
\end{figure}

Fig. \ref{fig:data-stats} provides more insights into the constitution of OmniNOCS. We use images from 10 other 3D detection datasets, which vary in terms of the number of instances, number of categories, and the scene complexity (measured by number of instances per image). While Hypersim provides the greatset number of objects, SUN-RGBD contributes the largest number of categories. The most complex scenes are from Hypersim (indoors) or KITTI and Waymo (outdoors). We show more examples for diverse object categories from the OmniNOCS dataset in Fig. \ref{fig:dataset-fig}. 


\subsection{Canonical orientation labeling}

\begin{figure}[thpb]
  \centering
  \includegraphics[width=0.98\linewidth]{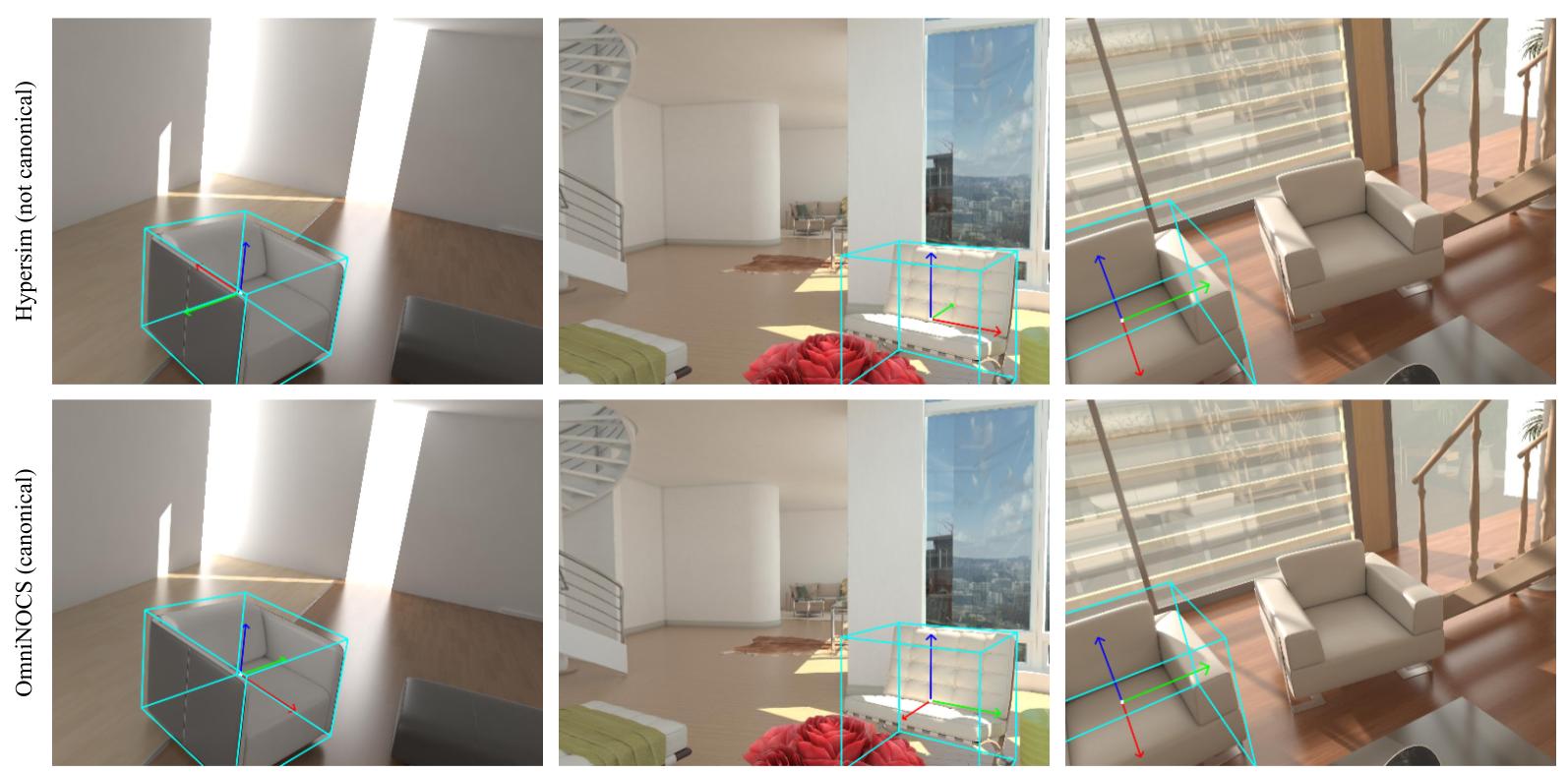}
   \caption{\small{\textbf{Instance-level orientation canonicalization for Hypersim}: Each chair instance in Hypersim has its X (red) and Y (green) axes chosen differently. To produce consistent NOCS coordinates, we apply an offset rotation to the original orientations such that the resulting orientations are consistent across all instances of the class. We find this offset by manually inspecting all Hypersim objects.}}
   \label{fig:suppl-hs-canon}
\end{figure}

To produce NOCS that are consistent across a category in OmniNOCS, it is required to have bounding boxes with consistently oriented axes. For example, all chairs in OmniNOCS have the Z-axis pointing upwards, and X axis pointing forwards. The definition of the canonical orientation is class-dependent. For some classes this is fully defined by geometry and the direction of gravity. For example, we use the longest edge of a book as its Z axis, and the shortest edge as the X axis. For windows, we use the upward direction as the Z axis and the shortest edge as the X axis, facing the direction of the camera. For other categories, such as chairs, desks, vehicles, pedestrians, these are defined based on the object semantics, and these need to be manually labeled. For example, most people would agree on what the front of a chair is, even though this might not be clear from the bounding box dimensions alone. We  choose to label this front direction as the X axis of the chair, and make sure to do so consistently across the entire dataset.

For the constituent datasets of Omni3D, the existing bounding box labelling is typically inconsistent between datasets (inter-dataset inconsistency). It can also be inconsistent within a dataset (intra-dataset inconsistency). We explain how we resolve this in each case below:


\textbf{Intra-dataset Inconsistency:}
This means that orientations for different instances of the same category can be inconsistent, within a single dataset. This happens in the case of Hypersim -- the orientation axes for a bounding box are chosen based on instance geometry, and can vary across instances of the same category (see Fig. \ref{fig:suppl-hs-canon}). For each instance, we manually choose an offset rotation that makes the resulting orientation consistent across the category, and with the rest of OmniNOCS. 

\textbf{Inter-dataset Inconsistency:}
In this more common case, orientations for all instances of a category are consistent, but only within the smaller dataset. For example, all chairs in Objectron and SUN-RGBD are oriented consistently within each dataset, but they are not consistent with each other. In this case, we apply the same offset rotation to all instances from a particular dataset to ensure that they are consistent with the rest of OmniNOCS.

\section{Model details}

We discuss the details of our implementation of {\nocsmethod} and {\boxmethod}, including the data processing, augmentation, model architecture details, and training regime in the following sections.

\subsection{Data augmentation}

We augment our training data with resizing by scaling the images randomly in the range [0.5, 1.5]. We also scale the camera intrinsics accordingly to ensure the 3D ground truth remains correct after augmentation. 

\subsection{Model architecture}

\textbf{Backbone:} We use the DINOv2 B-14 \cite{oquab2023dinov2} transformer as our image backbone, with a higher input image resolution of $896 \times 896$. We also use DPT layers \cite{ranftl2021vision} that fuse features from every third DINOv2 self-attention block, and upsamples the final features by a factor of 8 with a feature dimension of 512.

\noindent \textbf{\boxmethod{}}: While \nocsmethod{} regresses object coordinates, models like \cubercnn{} \cite{brazil2023omni3d} directly regress bounding box parameters: the projected 3D centroid, 3D size, 3DoF orientation, and depth. While a direct comparison of the localization accuracy of \nocsmethod{} to \cubercnn{} is provided in Table 4, this is not a fair comparison because \cubercnn{} uses a different convolutional backbone that is trained from scratch on the target dataset for both 2D region proposal and 3D localization. In contrast, \nocsmethod{} is a localization model that accepts 2D bounding box inputs. We therefore design a baseline we term \emph{\boxmethod{}}, that uses the same image backbone as \nocsmethod{}, input 2D bounding boxes, and predicts 3D bounding boxes using a self-attention based cube head unlike the convolutional head of \cite{brazil2023omni3d}. While \boxmethod{} can be supervised using Chamfer losses like \cubercnn{} \cite{brazil2023omni3d}, this causes the predicted orientations to be inconsistent. We therefore use direct L1 or L2 losses in the output parameter space (6D orientation vector, 2D projected centroid, scalar depth, and 3D size) as supervision. Our \boxmethod{} baseline therefore has a much better mAOE compared to \cubercnn{} in Table 4. However, our \nocsmethod{}'s orientations are  better than both \cubercnn{} and \boxmethod{}.

\subsection{Self-supervised reprojection error}
\label{sec:self-sup-supp}

As explained in Section 4.3, we use the NOCS reprojection error as self-supervision for training \nocsmethod{}. The loss is inspired from \cite{monorun2021} but unlike them, we do not require the uncertainty modelling / KL divergence step. We consider a 3D NOCS prediction $\mathbf{n}$ corresponding to a 2D pixel with image coordinates $\mathbf{p}$. The ground truth orientation is ${^c\mathbf{R}}_{o_{gt}}$, translation ${^c\mathbf{t}}_{o_{gt}}$ and scale $\mathbf{s}_{gt}$. We obtain 3D object coordinates, first in object frame, ${^o\mathbf{x}}$, then in camera frame ${^c\mathbf{x}}$, and then project to the image to obtain the reprojected NOCS point $\mathbf{p}_{proj}$

\begin{gather*}
    ^o\mathbf{x} = \mathbf{s}_{gt} . \mathbf{n} \\
    ^c\mathbf{x} =  {^c{\mathbf{R}}}_{o_{gt}}  {^o\mathbf{x}} + {^c\mathbf{t}}_{o_{gt}} \\
    \tilde{\mathbf{p}}_{proj} = \mathbf{K}_c  {^c\mathbf{x}}
\end{gather*}

\noindent where $.$ denotes element-wise scalar multiplication, $\mathbf{K}_c $ is the camera intrinsic matrix and $\tilde{\mathbf{p}}_{proj}$ is the homogeneous form of the projected point $\mathbf{p}_{proj}$. The self-supervised loss is given by: 

\begin{gather*}
\mathcal{L}_{ss} = 
\begin{cases}
    || \mathbf{p} - \mathbf{p}_{proj} ||_2, & \text{mask}_p = 1\\
    0,              & \text{otherwise}
\end{cases}
\end{gather*}

\noindent where $ \text{mask}_p$ is the predicted instance mask at $\mathbf{p}$. Note that we do not use gradients from this loss to supervise the predicted mask. The loss is termed self-supervised, as it does not need a ground truth NOCS map, but it requires the ground truth pose and size labels.

\subsection{Training}

We train our models using the Adam optimizer with a base learning rate of 1e-4 for 200k steps. We use a linear warmup of the learning rate over the first 1000 steps. We use a weight decay of 1e-6 for the weights of the convolutional layers and 1e-4 for the MLP layers. We use a dropout of 20 \% for the MLPs. We clip the global gradient norm to $10.0$. We use a batch size of 128, divided amongst 16 TPU cores (or 16 A100 GPUs). The models take approximately 40 hours to train. 

\subsection{Per-category NOCS quality}

Section 5.1 in the paper quantitatively evaluate the quality of NOCS produced by the model using the mAE, mPSNR and mIoU metrics. The numbers for these metrics in Table 3 are averaged over 75 classes in OmniNOCS that have high quality NOCS ground truth. Here, we provide the per-class numbers in Table \ref{table:suppl-per-class-nocs} to analyze the variability of predictions among classes.  In general, we see that categories that are either rare or very diverse have higher NOCS errors compared to other categories. For example, toys and projectors have PSNR of 17.92 and 12.21 respectively, whereas cars have a PSNR of 27.45.

\section{Temporal consistency results}

We provide a video of NOCSformer's independent predictions on each frame for some sequences from Objectron \cite{objectron2021}, attached in the supplementary material. The independent NOCS and pose estimates are consistent temporally, without use of any smoothing/filtering techniques. 

\section{Discussion and limitations}

\nocsmethod{} predicts 3D object coordinates aligned to 2D pixel values, yielding 3D-2D correspondences that can be used to estimate the 6DoF object pose. It shows that models that predict 3D-2D correspondences can be scaled to larger datasets and diverse classes, enabling widespread adoptability. This is an alternative to directly regressing the object pose from an image. \nocsmethod{} exhibits both pros and cons compared to methods that directly regress object poses. 

\textbf{More flexible representation:} Predicting NOCS allows for different methods to be used for estimating the object pose, based on the application. The options are 1) using learned network heads to predict pose from NOCS (as in NOCSformer) 2) using PnP variants to estimate pose from 3D-2D correspondences 3) using 3D-3D alignment to estimate pose, if a depth sensor is available.

\textbf{More interpretable:} Predicting pose from 3D-2D correspondences is more interpretable than using an end-to-end trained model. 


\textbf{Less accurate at longer ranges:} For small objects at very long ranges (such as those in outdoor self-driving scenes), the accuracy of \nocsmethod{} deteriorates. Since the RoI resolution is higher than the size of these objects, the input to the RoI heads is itself coarse and less informative, producing more noisy masks, NOCS and size predictions. The depth estimates at longer ranges are more sensitive to errors in object coordinate predictions, causing higher depth errors. For long ranges, particularly in single-camera applications, it may be more accurate to regress pose directly from an end-to-end model as they rely on dataset biases.

\begin{table}[tb]
\centering

\resizebox{\linewidth}{!}{
\begin{tabular}{l|ccccccccccccccccccc}
\hline
Class & car   & blinds & van   & monitor & curtain & mirror & toilet & air conditioner & closet & stove & cyclist & board & clothes & pedestrian & toaster & dresser & painting & bookcase & shelves \\
 \hline
MAE & 0.037 & 0.038  & 0.041 & 0.041   & 0.045   & 0.047  & 0.047  & 0.05            & 0.054  & 0.056 & 0.056   & 0.057 & 0.058   & 0.058      & 0.059   & 0.061   & 0.061    & 0.062    & 0.062   \\
PSNR & 27.45 & 27.92  & 26.07 & 26.77   & 26.14   & 25.97  & 24.31  & 25.97           & 22.68  & 22.86 & 22.98   & 26.0  & 21.19   & 23.8       & 22.45   & 23.74   & 23.53    & 23.03    & 23.24   \\
IoU & 0.84  & 0.943  & 0.831 & 0.814   & 0.874   & 0.758  & 0.953  & 0.845           & 0.974  & 0.95  & -     & 0.937 & 0.227   & 0.659      & 0.958   & 0.908   & 0.884    & 0.853    & 0.675   \\
\hline
\end{tabular}
}
\resizebox{\linewidth}{!}{
\begin{tabular}{l|ccccccccccccccccccc}
\hline
Class & door  & truck & bottle & printer & sofa  & computer & fireplace & picture & chair & vase  & person & coffee maker & sink  & potted plant & drawers & television & bed   & keyboard & microwave \\
\hline
MAE & 0.064 & 0.065 & 0.066  & 0.066   & 0.067 & 0.068    & 0.069     & 0.071   & 0.071 & 0.071 & 0.073  & 0.075        & 0.076 & 0.076        & 0.079   & 0.079      & 0.08  & 0.08     & 0.084     \\
PSNR & 23.87 & 22.7  & 22.42  & 21.12   & 22.88 & 22.55    & 21.99     & 22.19   & 22.07 & 21.29 & 21.69  & 20.03        & 20.62 & 21.61        & 21.47   & 21.37      & 21.32 & 21.64    & 18.65     \\
IoU & 0.856 & 0.843 & 0.856  & 0.776   & 0.791 & 0.935    & 0.928     & 0.92    & 0.724 & 0.787 & 0.498  & 0.933        & 0.858 & 0.86         & 0.87    & 0.929      & 0.599 & 0.855    & 0.778     \\
\hline

\end{tabular}
}
\resizebox{\linewidth}{!}{
\begin{tabular}{l|ccccccccccccccccccc}
\hline
Class & machine & plates & lamp  & soundsystem & cup   & cabinet & refrigerator & night stand & bathtub & fan   & rack  & tray  & towel & tissues & pen   & bowl  & oven  & desk  & phone \\
\hline
MAE & 0.085   & 0.089  & 0.089 & 0.09        & 0.09  & 0.091   & 0.092        & 0.093       & 0.094   & 0.097 & 0.097 & 0.098 & 0.098 & 0.1     & 0.101 & 0.106 & 0.106 & 0.11  & 0.114 \\
PSNR & 20.44   & 19.31  & 20.5  & 19.62       & 19.32 & 21.49   & 19.17        & 20.02       & 19.73   & 18.78 & 18.83 & 18.34 & 19.23 & 18.77   & 18.69 & 18.39 & 18.16 & 18.85 & 17.73 \\
IoU & 0.721   & 0.953  & 0.676 & 0.899       & 0.819 & 0.879   & 0.9          & 0.794       & 0.958   & 0.386 & 0.651 & 0.657 & 0.951 & 0.769   & 0.688 & 0.86  & 0.914 & 0.76  & 0.876 \\
\hline

\end{tabular}
}

\resizebox{\linewidth}{!}{
\begin{tabular}{l|cccccccccccccccccc}
\hline
Class & bag   & toys  & table & blanket & bin   & cart  & fire extinguisher & faucet & kitchen pan & pillow & stationery & box   & utensils & counter & books & tram  & electronics & projector \\
\hline
MAE & 0.116 & 0.116 & 0.117 & 0.119   & 0.119 & 0.12  & 0.122             & 0.124  & 0.128       & 0.129  & 0.131      & 0.138 & 0.148    & 0.15    & 0.153 & 0.169 & 0.176       & 0.185     \\
PSNR & 17.58 & 17.92 & 18.18 & 16.69   & 17.81 & 16.42 & 15.06             & 16.8   & 16.46       & 17.1   & 16.49      & 16.6  & 14.99    & 15.95   & 15.23 & 14.06 & 13.8        & 12.21     \\
IoU & 0.873 & 0.553 & 0.682 & 0.945   & 0.86  & 0.852 & 0.901             & 0.774  & 0.867       & 0.728  & 0.778      & 0.842 & 0.286    & 0.705   & 0.773 & -   & 0.883       & 0.933     \\
\hline

\end{tabular}
}
\vspace{2mm}
\caption{\small{\textbf{Quality of NOCS predictions per class:} As discussed in Section 5.1 of the paper, we adopt the use of mAE, mPSNR, and mIoU to measure the quality of object NOCS produced by a model on OmniNOCS test set. We provide the split for these numbers per class here. The mIoU for some categories are not available as these do not have ground truth mask annotations in the test set.}}
\label{table:suppl-per-class-nocs}
\end{table}

\bibliographystyle{splncs04}
\bibliography{main}
\end{document}